\documentclass[sigconf]{acmart}
\AtBeginDocument{%
  }

\AtBeginDocument{%
  }

\copyrightyear{2026}
\acmYear{2026}
\setcopyright{cc}
\setcctype{by-nc-nd}
\acmConference[CIKM '26] {Proceedings of the 35th ACM International Conference on Information and Knowledge Management}{November 7--11, 2026}{Rome, Italy.}
\acmBooktitle{Proceedings of the 35th ACM International Conference on Information and Knowledge Management (CIKM '26), November 7--11, 2026, Rome, Italy}
\acmISBN{979-8-4007-2539-5/2026/11}
\acmDOI{10.1145/3799682.3840176}

\usepackage{amsmath,amsfonts,bm}

\def\eqref#1{equation~\ref{#1}}
\def\1{\bm{1}}

\DeclareMathAlphabet{\mathsfit}{\encodingdefault}{\sfdefault}{m}{sl}
\SetMathAlphabet{\mathsfit}{bold}{\encodingdefault}{\sfdefault}{bx}{n}

\usepackage{hyperref}
\usepackage{url}
\usepackage{booktabs}       % professional-quality tables
\usepackage{amsfonts}       % blackboard math symbols
\usepackage{nicefrac}       % compact symbols for 1/2, etc.
\usepackage{xr}
\usepackage{microtype}      % microtypography
\usepackage{soul}         % colors
\usepackage{amsmath,bm}
\usepackage{multirow}
\usepackage{array}
\usepackage{makecell}
\usepackage{graphicx}
\usepackage{diagbox}
\usepackage{tabularx}
\usepackage{adjustbox}
\usepackage{bbding}
\usepackage{transparent}
\usepackage{subcaption}
\usepackage{amsfonts}
\usepackage{enumitem}
\usepackage{xspace}
\usepackage{tablefootnote}
\usepackage{xcolor}

\newcommand{\nop}[1]{}

\newcommand{\data}{LEMMA-RCA}

\usepackage[noabbrev,nameinlink]{cleveref}
\newcolumntype{x}[1]{>{\centering\arraybackslash\hspace{0pt}}p{#1}}

\makeatletter
\newcommand{\thickhline}{%
    \noalign {\ifnum 0=`}\fi \hrule height 1.2pt
    \futurelet \reserved@a \@xhline
}

\newcommand{\rca}{\ensuremath{\textnormal{\text{LEMMA-RCA}}}\xspace}

\title{LEMMA-RCA: A Large Multi-modal Multi-domain Dataset for Root Cause Analysis}

\author{Lecheng Zheng}
\authornote{Equal Contribution.}
\email{lecheng4@illinois.edu}
\affiliation{%
  \institution{University of Illinois Urbana-Champaign}
  \country{USA}
}

\author{Zhengzhang Chen$^*$}
\email{zchen@nec-labs.com}
\affiliation{%
  \institution{NEC Laboratories America}
  \country{USA}
}

\author{Dongjie Wang$^*$}
\email{wangdongjie100@gmail.com}
\affiliation{%
  \institution{University of Kansas}
  \country{USA}
}
\author{Chengyuan Deng$^*$}
\email{cd751@rutgers.edu}
\affiliation{%
  \institution{Rutgers University}
  \country{USA}
}
\author{Reon Matsuoka}
\email{r-matsuoka@nec.com}
\affiliation{%
  \institution{NEC Laboratories Japan}
  \country{Japan}
}
\author{Haifeng Chen}
\email{Haifeng@nec-labs.com}
\affiliation{%
  \institution{NEC Laboratories America}
  \country{USA}
}

\renewcommand{\shortauthors}{Lecheng Zheng et al.}
\begin{document}

\begin{abstract}
Root cause analysis (RCA) is crucial for enhancing the reliability and performance of complex systems. However, progress in this field has been hindered by the lack of large-scale, open-source datasets tailored for RCA. To bridge this gap, we introduce LEMMA-RCA, a large dataset designed for diverse RCA tasks across multiple domains and modalities. LEMMA-RCA features various real-world fault scenarios from Information Technology (IT) and Operational Technology (OT) systems, including microservices, water distribution, and water treatment systems, with hundreds of system entities involved. We evaluate the performance of six baseline methods on LEMMA-RCA across various settings, including offline and online models, as well as single and multi-modal settings. Our study demonstrates the utility of LEMMA-RCA in facilitating fair evaluation and promoting the development of more robust RCA techniques.
The dataset is publicly available at \url{https://lemma-rca.github.io/}, with archival metadata and access instructions deposited under DOI: \url{https://doi.org/10.5281/zenodo.20516735}.
\end{abstract}

\maketitle

\section{Introduction}

Root cause analysis (RCA) is essential for identifying the underlying causes of system failures and ensuring the reliability of real-world systems~\citep{fourlas2021survey, 7069265, DBLP:conf/aistats/PamfilSDPGBA20, DBLP:conf/nips/NgG020}. As modern systems grow in complexity and interdependence, faults arising from interactions among modular services can disrupt user experiences and incur significant financial losses. Manual RCA is labor-intensive, costly, and error-prone, making efficient data-driven RCA methods crucial for fault localization and loss mitigation.

RCA has been extensively studied across various domains and settings~\citep{capozzoli2015fault,deng2021graph,brandon2020graph}, and can be conducted in an \emph{offline/online} fashion with \emph{single/multi-modal} data. Existing approaches leverage techniques such as Bayesian methods~\citep{alaeddini2011using} and decision trees~\citep{chen2004failure}. In particular, causal structure learning~\citep{DBLP:journals/technometrics/Burr03,DBLP:journals/pami/TankCFSF22,yu2023nezha,wang2023incremental,wang2023interdependent,zheng2024mulan} has proven effective for constructing causal or dependency graphs between system entities and key performance indicators (KPIs).

Despite progress in RCA techniques, large-scale public datasets remain scarce due to confidentiality concerns~\citep{harsh2023murphy}, hindering fair comparison. Available datasets typically contain manually injected faults and cover only a single domain, limiting their utility for real-world scenarios and potentially leading to regulatory and ethical consequences in critical sectors.

To address these limitations, we introduce \textbf{LEMMA-RCA}, a collection of \underline{L}arg\underline{E}-scale \underline{M}ulti-\underline{M}od\underline{A}l datasets with real system faults to facilitate future \underline{R}oot \underline{C}ause \underline{A}nalysis research. LEMMA-RCA spans real-world applications including IT operations and water treatment systems with \textbf{hundreds of system entities}, and accommodates \textbf{multi-modal} data including textual system logs with millions of event records and time series metrics with over 100,000 timestamps. We annotate ground-truth labels marking the precise timestamps of \textbf{real system faults} and their root-cause entities.

\Cref{table:existing_dataset} compares LEMMA-RCA with existing datasets: \textit{NeZha}~\citep{yu2023nezha} is limited in size, has missing monitoring data, and is confined to microservices; \textit{PetShop}~\citep{pet_shop} is small with only 41 components/pods, limiting real-world practicality; \textit{Sock-Shop}~\citep{ikram2022root} contains only 13 microservices with synthetic faults, is not public, and is single-modal; \textit{ITOps}~\citep{li2022constructing} is non-public and offers only structured logs unsuitable for fine-grained RCA; \textit{Murphy}~\citep{harsh2023murphy} is collected from a simple system and is also non-public. In contrast, LEMMA-RCA offers comprehensive maturity in accessibility, authenticity, and diversity. For broader impact, ethical considerations, and limitations, please refer to Appendix \ref{apdx:disc}.

LEMMA-RCA enables fair comparison across RCA methods. We evaluate six baselines and assess data modality quality in offline settings, then adapt these methods for online evaluation under a standardized LEMMA-RCA formulation, enabling, for the first time, rigorous comparison of online RCA methods on common ground. Experimental results demonstrate LEMMA-RCA's effectiveness and broad utility for advancing RCA research.

\begin{table}[tp]
\tiny
\centering
\renewcommand{\arraystretch}{1}
\caption{\textbf{Existing datasets for root cause analysis.} The top row corresponds to our dataset. The
symbols \Checkmark and {\XSolidBrush} indicate that the dataset has or does not have the corresponding feature, respectively.} 
\vspace{-2mm}
\label{table:existing_dataset}
\begin{adjustbox}{width=\linewidth}
\begin{tabular}{c|c|c|c|c|c|c} 
\thickhline
\multirow{2}{*}{Dataset}    & \multirow{2}{*}{Public} & \multirow{2}{*}{Real Faults}  & \multirow{2}{*}{Large-scale} & \multirow{2}{*}{Multi-domain}   & \multicolumn{2}{c}{Modality} \\ 
\cline{6-7}
&   &   &  &   & Single & Multiple     \\ 
\hline
\rca        & \Checkmark               & \Checkmark           & \Checkmark              & \Checkmark    &  \Checkmark &\Checkmark\\
 NeZha & \Checkmark               & \transparent{0.2}{\XSolidBrush}           & \transparent{0.2}{\XSolidBrush}    & \transparent{0.2}{\XSolidBrush}  & \Checkmark        & \Checkmark  \\
 PetShop & \Checkmark                & \transparent{0.2}{\XSolidBrush}           & \transparent{0.2}{\XSolidBrush}   & \transparent{0.2}{\XSolidBrush} & \Checkmark          & \transparent{0.2}{\XSolidBrush} \\
       Sock-Shop & \transparent{0.2}{\XSolidBrush}                  & \transparent{0.2}{\XSolidBrush}            & \transparent{0.2}{\XSolidBrush} & \transparent{0.2}{\XSolidBrush}    & \Checkmark        & \transparent{0.2}{\XSolidBrush} \\
  ITOps & \transparent{0.2}{\XSolidBrush}                 & \Checkmark           & \Checkmark     & \transparent{0.2}{\XSolidBrush} & \Checkmark        & \transparent{0.2}{\XSolidBrush} \\
   Murphy & \transparent{0.2}{\XSolidBrush}                 & \Checkmark           & \transparent{0.2}{\XSolidBrush}  & \transparent{0.2}{\XSolidBrush}    & \Checkmark        & \transparent{0.2}{\XSolidBrush} \\
\thickhline
\vspace{-6mm}
\end{tabular}
\end{adjustbox}
\end{table}

\section{Preliminaries and Related Work}
\begin{figure}[h]
    \centering
    \includegraphics[width=\linewidth]{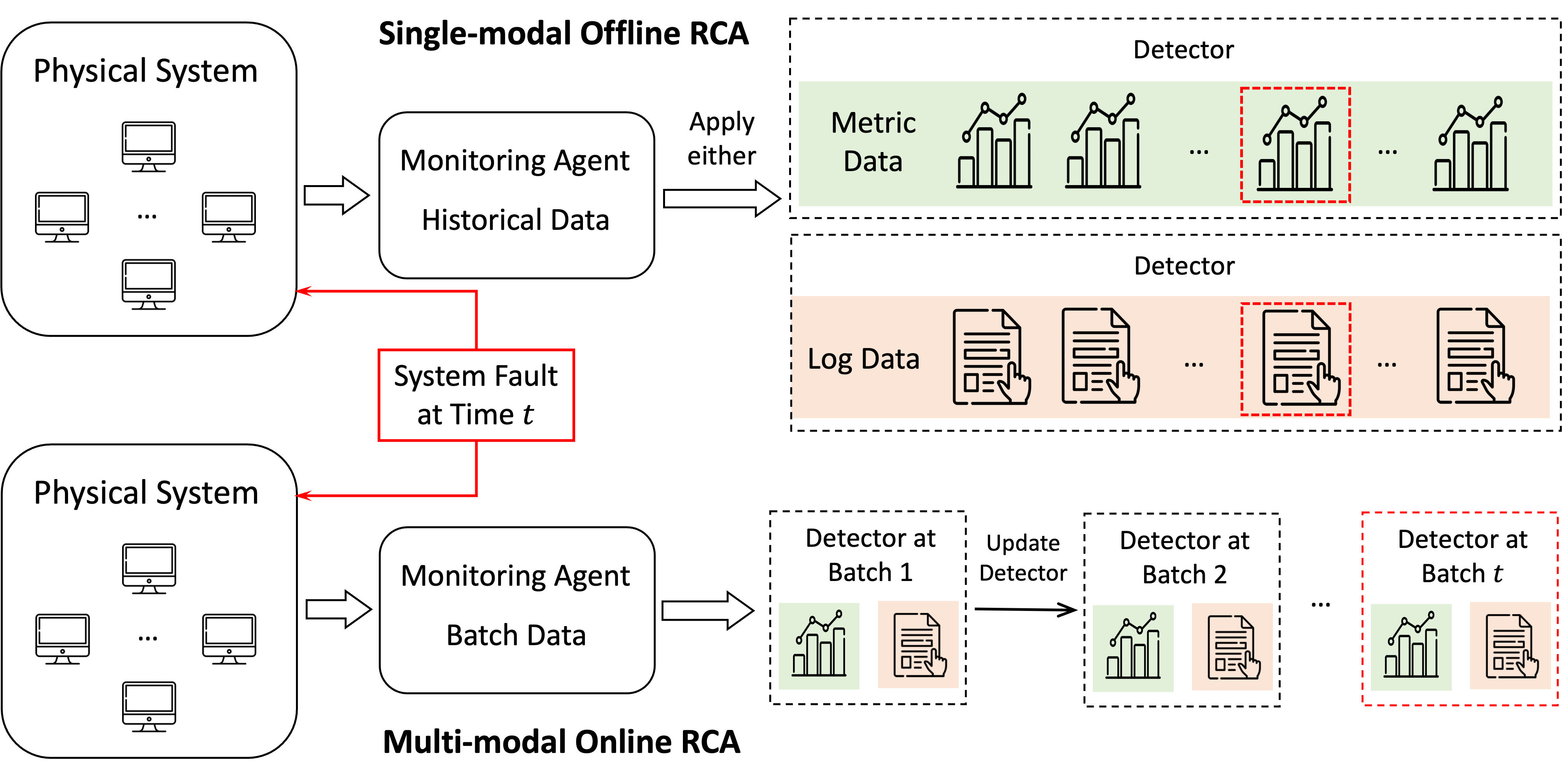}
     \vspace{-2mm}
    \caption{Illustration of RCA workflow in the single-modal offline setting (top) and the multi-modal online setting (bottom). The other two settings can be viewed as an ensemble of corresponding components (data collection, detector, modality) and follow the same systematic procedure.}
    \label{fig:rca-intro}
\end{figure}

\noindent\textbf{Key Performance Indicator (KPI)} is a monitoring time series that reflects system status (e.g., latency, response time in microservices), where abnormal values often signal performance degradation or failure.\\
\noindent\textbf{Entity Metrics} are multivariate time series collected from system components (e.g., machines, containers, pods), such as CPU, memory, and disk IO utilization. Abnormal entities are often potential root causes of failures. \\
\noindent\textbf{Data-driven Root Cause Analysis Problem.}
Given monitoring data (metrics and logs) of system entities and KPIs, RCA aims to identify the top-K entities most relevant to KPIs when a fault occurs. RCA settings are commonly categorized along two dimensions: offline (retrospective, historical data) vs.\ online (real-time streams), and single-modal vs.\ multi-modal. \Cref{fig:rca-intro} illustrates the single-modal offline and multi-modal online procedures.\\
\noindent\textbf{Single-modal Offline RCA} identifies fault causes retrospectively from a single data type~\citep{wang2023interdependent,tang2019graph,meng2020localizing,li2021practical,soldani2022anomaly}. Representative works include sequential and causal-temporal modeling of metrics~\citep{meng2020localizing}, random walks on interdependent causal networks~\citep{wang2023interdependent}, and trace-based localization via abnormal/normal ratios~\citep{li2021practical}. LLM-based approaches have also emerged as a promising direction for learning causal relations~\citep{chen2024automatic, shan2024face, goel2024x, zhou2024causalkgpt, roy2024exploring, wang2023rcagent}. However, reliance on a single modality may yield biased or suboptimal results.\\
\noindent\textbf{Multi-modal Offline RCA} follows two main paradigms~\citep{yu2023nezha, DBLP:conf/ispa/HouJWLH21, zheng2024mulan, lan2023mm}: (i) extracting information per modality and then integrating it, as in Nezha~\citep{yu2023nezha} and PDiagnose~\citep{DBLP:conf/ispa/HouJWLH21}; and (ii) modeling cross-modal interactions, such as MULAN~\citep{zheng2024mulan}, which learns inter-modal correlations to build a unified causal graph, and MM-DAG~\citep{lan2023mm}, which jointly learns multiple DAGs.\\
\noindent\textbf{Online RCA.} Most RCA methods operate offline, requiring full retraining for new faults and delaying response. Wang \textit{et al.}~\citep{wang2023incremental} decouple state-invariant and state-dependent information to incrementally update the causal graph, while Li \textit{et al.}~\citep{DBLP:conf/kdd/0005LYNZSP22} leverage system architecture in a causal Bayesian network to reduce bias toward new data. Yet these methods remain single-modal, leaving online multi-modal RCA an open challenge.

\section{LEMMA-RCA Data}
 
This section outlines the data resources, preprocessing steps, and key characteristics of the released data. Detailed information on all fault scenarios, additional preprocessing steps for SWaT/WADI, and extended visualizations can be found in the supplementary materials (Appendix \ref{dataset_details}). The data license is provided in supplementary materials (Appendix~\ref{apx:license}).
 
\subsection{Data Collection}
\begin{figure*}[h]
\begin{center}
\includegraphics[width=0.9\linewidth]{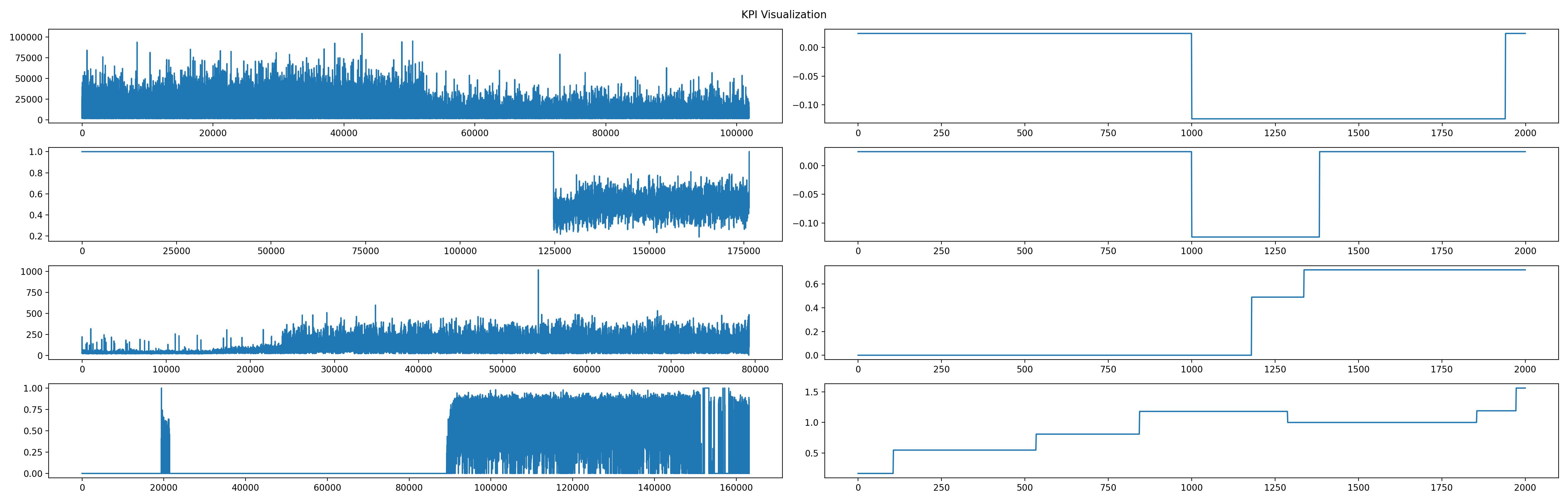}
\end{center}
\vspace{-1mm}
\caption{KPI visualization for system failures. \textbf{Left}: Product Review (first two) and Cloud Computing (last two); \textbf{Right}: SWaT (first two) and WADI (last two).}
\label{fig_aiops}
\end{figure*}

We collect real-world data from two domains: IT operations and OT operations. The IT domain includes \textbf{Product Review} and \textbf{Cloud Computing} microservice sub-datasets, while the OT domain includes \textbf{SWaT} and \textbf{WADI} sub-datasets from water treatment and distribution systems. Statistics are summarized in~\Cref{table_aiops_statistics}.
The \textbf{Product Review} Platform comprises six OpenShift nodes and $216$ system pods, collecting four fault types (out-of-memory, high-CPU-usage, external-storage-full, and DDoS attack), each lasting at least $49$ hours. Node- and pod-level metrics were recorded by Prometheus~\citep{turnbull2018monitoring} at 1-second granularity, while logs were collected by ElasticSearch~\citep{zamfir2019systems}. JMeter~\citep{nevedrov2006using} tracked system status, with latency serving as the KPI. The \textbf{Cloud Computing} Platform captures six fault types (e.g., cryptojacking, GitOps mistakes, configuration change failure), with metrics from CloudWatch Metrics and logs (general, API debug, MySQL) from CloudWatch Logs; latency, error rate, and utilization rate serve as KPIs.
In the OT domain, \textbf{SWaT} and \textbf{WADI}~\citep{swat_wadi} provide time-series sensor/actuator data from testbeds at iTrust lab. SWaT~\citep{mathur2016swat} covers $11$ days with $51$ sensors and $16$ faults, while WADI~\citep{ahmed2017wadi} covers $16$ days with $123$ sensors/actuators and $15$ faults. \Cref{fig_aiops} visualizes KPIs for eight failure cases, where sudden spikes or drops in latency indicate system failures across all four sub-datasets.
 
\begin{table}[tp]
\vspace{-2mm}
\renewcommand\arraystretch{1.1}
\scriptsize
\centering
\caption{Data statistics of IT and OT operation sub-datasets.}
\vspace{-2mm}
\begin{adjustbox}{width=\linewidth}
\begin{tabular}{ccc}
\thickhline
Microservice System (IT) & Product Review & Cloud Computing \\ \hline
Original Dataset Size & 765 GB & 540 GB \\
Number of (\#) fault types & 4 & 6 \\
Average \# entities per fault & 216.0 & 167.71 \\
Average \# metrics per fault & 11 (node) + 6 (pod) & 6 (node) + 7 (pod) \\
Average \# timestamps per fault & 131{,}329.25 & 109{,}350.57 \\
Average max log events per fault & 153{,}081{,}219.0 & 63{,}768{,}587.25 \\
\thickhline
Water Treatment/Distribution (OT) & SWaT & WADI \\ \hline
Original Dataset Size & 4.47G & 5.67G \\
Number of (\#) fault types & 16 & 9 \\
Average \# entities per fault & 51.0 & 123.0 \\
Average \# metrics per fault & 7 (node) + 7 (pod) & 7 (node) + 7 (pod) \\
Average \# timestamps per fault & 56{,}239.88 & 85{,}248.47 \\
\thickhline
% \vspace{-8mm}
\end{tabular}
\end{adjustbox}
\label{table_aiops_statistics}
\end{table}

\subsection{Data Preprocessing}
 
After collection, we exclude non-stationary pods to retain only those suitable for modeling. For \textbf{Product Review} and \textbf{Cloud Computing}, following~\citep{zheng2025online}, we transform unstructured logs into time-series via three feature types: (i) $X^L_1 \in \mathbb{R}^{T}$, the occurrence frequency of structured log templates extracted by Drain over 10-minute windows with 30-second intervals; (ii) $X^L_2 \in \mathbb{R}^{T}$, the frequency of abnormal templates containing golden-signal keywords (e.g., `error', `exception', `critical'); and (iii) $X^L_3 \in \mathbb{R}^{T}$, the leading PCA component of TF-IDF features. The final feature matrix is $X^L = [X^L_1; X^L_2; X^L_3] \in \mathbb{R}^{3 \times T}$. For \textbf{SWaT} and \textbf{WADI}, since the original label column is discrete, we apply anomaly detection algorithms (e.g., Support Vector Data Description, Isolation Forest) to derive continuous anomaly scores serving as KPIs. The segmentation procedure for SWaT and WADI is described in Appendix \ref{ape:time}.

\subsection{System Fault Scenarios}
 
LEMMA-RCA covers $10$ distinct real-world fault types across Product Review and Cloud Computing. Due to the space limitation, we provide the details in online supplementary materials (Appendix~\ref{system_faul_cases}).

\section{Experiments}
\subsection{Experimental Setup}

\noindent \textbf{Evaluation Metrics}. To assess baseline RCA method on LEMMA-RCA, we choose three widely-used metrics~\citep{wang2023interdependent, DBLP:conf/iwqos/MengZSZHZJWP20, zheng2024mulan} and introduce them below.

\noindent (1). \textbf{Precision@K (PR@K)}: It measures the probability that the top $K$ predicted root causes are real, defined as: 
\begin{equation}
    \text{PR@K} = \frac{1}{|\mathbb{A}|}\sum_{a \in \mathbb{A}}\frac{\sum_{i<k}R_a(i)\in V_a}{\min (K, |v_a|)}
\end{equation}
where $\mathbb{A}$ is the set of system faults, $a$ is one fault in $\mathbb{A}$, $V_a$ is the real root causes of $a$, $R_a$ is the predicted root causes of $a$, and $i$ is the $i$-th predicted cause of $R_a$.

\noindent(2). \textbf{Mean Average Precision@K (MAP@K)}: It assesses the top $K$ predicted causes from the overall perspective, defined as:
\begin{equation}
    \text{MAP@K} = \frac{1}{K|\mathbb{A}|} \sum_{a \in \mathbb{A}} \sum_{i\leq j\leq K} \text{PR@j}
\end{equation}
where a higher value indicates better performance.

\noindent(3). \textbf{Mean Reciprocal Rank (MRR)}: It evaluates the ranking capability of models, defined as:
\begin{equation}
    \text{MRR@K} = \frac{1}{|\mathbb{A}|}\sum_{a \in \mathbb{A}}\frac{1}{\text{rank}_{R_a}}
\end{equation}
where $\text{rank}_{R_a}$ is the rank number of the first correctly predicted root cause for system fault $a$.

\noindent \textbf{Baselines}.
We evaluate the performance of the following RCA baselines on the benchmark sub-datasets, selecting only those with publicly available code to ensure fair and reproducible comparisons: (1) \textbf{PC-based}~\citep{ma2020automap}: 
This approach first employs a Peter-Clark (PC) algorithm~\cite{DBLP:journals/technometrics/Burr03} to construct the anomaly behavior graph, and then applies a random walk algorithm to rank the root causes based on the estimated graph structure. 
(2) \textbf{CIRCA}~\citep{li2022causal}: This model utilizes structural graph construction, regression-based hypothesis testing, and descendant adjustment to identify root cause metrics.
(3) \textbf{$\epsilon$-Diagnosis}~\citep{shan2019diagnosis}: This model diagnoses small-window, long-tail latency in large-scale microservice platforms using a two-sample test and $\epsilon$-statistics.
(4) \textbf{RCD}~\citep{ikram2022root}: This technique hierarchically localizes the root cause of failures by focusing on relevant sections of the dependency graph.
(5) \textbf{BARO}~\citep{pham2024baro}: It is an end-to-end approach integrating Bayesian change point detection and nonparametric hypothesis testing to detect anomalies and identify root causes in microservice systems.
(6) \textbf{Nezha}~\citep{yu2023nezha}: A multi-modal method designed to identify root causes by detecting abnormal patterns. 

When using only metric data (\textit{e.g.}, CPU usage, memory usage) for root cause analysis, we first identify the potential root cause associated with each metric. We then compute the final score for each candidate by averaging its ranking scores across all metrics. This procedure is similarly applied in log-only and multi-modal scenarios to rank and identify the root causes.
For the hyperparameters, we use the default parameter values for all baselines to ensure a fair comparison.
Reproducibility details, including hardware and software environments, are provided in Appendix \ref{ape:rep}.
Detailed descriptions of the evaluated baseline methods are provided in Appendix \ref{apx:baseline}.

\begin{table}[ht]
\vspace{-5pt}
\renewcommand{\arraystretch}{1.2}
\caption{Results for RCA baselines with multiple modalities on the Product Review dataset.}
\vspace{-2mm}
\small
\centering
\begin{adjustbox}{width=\linewidth}
\begin{tabular}{*{9}{c}}
\thickhline      Modality      & Model         & PR@1      & PR@5      & PR@10     & MRR       & MAP@3     & MAP@5     & MAP@10 \\ \cline{1-9} \thickhline
\multirow{6}{*}{Metric Only}  & PC           & 0         & 0         & 0.250      & 0.053     & 0         & 0         & 0.050  \\     
                             % & \textcolor{red}{PCMCI}           & \textcolor{red}{0.250}         & \textcolor{red}{0.500}         & \textcolor{red}{0.500}      & \textcolor{red}{0.342}     & \textcolor{red}{0.250}         & \textcolor{red}{0.300}         & \textcolor{red}{0.400}  \\      
                             & RCD       & 0  &  0.250  &  0.750  &  0.185  &  0.167  &  0.200  &  0.350  \\
                             & $\epsilon$-Diagnosis       & 0  &  0  &  0.250  &  0.038  &  0  &  0  &  0.050  \\
                             & CIRCA       & 0  &  0.750  &  0.750  &  0.283  &  0.250  &  0.450  &  0.600  \\
                            %  & Granger       & 0.250    & 0.750    & 0.750      & 0.474    & 0.500    & 0.250    & 0.675  \\
                             & BARO       & 0.250  &  0.250  &  0.250  &  0.286  &  0.250  &  0.250  &  0.250 \\
                             \hline  
\multirow{6}{*}{Log Only}    & PC           & 0         & 0         & 0.250      & 0.069     & 0         & 0         & 0.125 \\ 
                             & RCD       & 0  &  0  &  0.250  &  0.056  &  0  &  0  &  0.025  \\
                             & $\epsilon$-Diagnosis     & 0  &  0  &  0.250  &  0.038  &  0  &  0  &  0.050  \\
                             & CIRCA & 0  &  0  &  0.500  &  0.080  &  0  &  0  &  0.125  \\
                             % & Granger       & 0         & 0         & 0.250      & 0.0590     & 0         & 0         & 0.075  \\ 
                             & BARO & 0      & 0.250       & 0.250      & 0.139     & 0.167     & 0.200      & 0.225  \\
                             \hline
\multirow{7}{*}{Multi-Modality} & PC                    &  0      & 0      & 0.250   & 0.064  & 0      & 0      & 0.125 \\  
                                & RCD	                &  0  &  0.500  &  0.750  &  0.231  &  0.167  &  0.300  &  0.525 \\
                                & $\epsilon$-Diagnosis	&  0  &  0  &  0.250  &  0.041  &  0  &  0  &  0.075 \\
                                & CIRCA	                &  0  &  \textbf{0.750}  &  \textbf{1.000}  &  0.299  &  0.250  &  0.450  &  0.650 \\
                                & BARO                  &  \textbf{0.750}  &  \textbf{0.750}  &  \textbf{1.000}  &  \textbf{0.775}  &  \textbf{0.750}  &  \textbf{0.750}  &  \textbf{0.775} \\
                                & Nezha	                &  0  & 0.500  & 0.750  & 0.193  & 0.083  & 0.250  & 0.475 \\
\thickhline
\end{tabular}
\end{adjustbox}
\label{table_result_1}
\vspace{-5pt}
\end{table} 

\subsection{Root Cause Analysis Results}

\noindent \textbf{Product Review and Cloud Computing}.
We evaluate six RCA methods including both single-modal and multi-modal methods on Product Review and Cloud Computing sub-datasets. The experimental results are presented in~\Cref{table_result_1} with respect to Precision at K (PR@K), Mean Reciprocal Rank (MRR), and Mean Average Precision at K (MAP@K).
Due to space limitations, the experimental results on Cloud Computing (Table~\ref{table_result_2}) are reported in the Appendix~\ref{more_results}. 
Our observations reveal the following insights:  
(1) The PC algorithm and $\epsilon$-Diagnosis perform worst on both the Product Review and Cloud Computing sub-datasets. We conjecture that PC, RCD, and $\epsilon$-Diagnosis struggle to capture long-term dependencies in large-scale datasets, making it difficult to detect abnormal temporal patterns.  
(2) CIRCA outperforms RCD and $\epsilon$-Diagnosis, consistent with findings from the Petshop study~\citep{pet_shop}, where CIRCA's regression-based hypothesis testing and adjustment mechanisms led to higher diagnostic accuracy.  
(3) Multi-modal input—combining both metric and log data—significantly enhances the performance of RCA methods compared to using either modality alone. For example, BARO achieves only 25\% PR@1 with metric data and 0\% with log data on the Product Review sub-dataset, but reaches 75\% PR@1 when both are used, correctly identifying the root cause in 75\% of fault scenarios. This highlights the complementary nature of log and metric data and the importance of integrating both to improve diagnostic accuracy and overall performance, particularly in terms of MRR.

\begin{table}[!ht]
\renewcommand\arraystretch{1.2}
\small
\caption{Results for RCA baselines on the SWaT sub-dataset.} 
\vspace{-2mm}
\centering
\begin{adjustbox}{width=\linewidth}
\begin{tabular}{ccccccccc}
\thickhline
 Dataset & Model  & PR@1 & PR@5 & PR@10 & MRR & MAP@3 & MAP@5 & MAP@10 \\ \thickhline
 
\multirow{7}{*}{SWaT} 
 & PC & 0.125 & \textbf{0.344} & 0.583 & 0.262 & 0.129 & \textbf{0.204} & 0.350        \\
 & RCD       & 0.125        & 0.125        & 0.625       & 0.228       & 0.125     & 0.125      & 0.344  \\
& $\varepsilon$-Diagnosis       & 0.125        & 0.125        & 0.563       & 0.217     & 0.125    & 0.125      & 0.294  \\
& CIRCA       & \textbf{0.188}      & 0.250       & \textbf{0.688}       & \textbf{0.287}       & \textbf{0.188}     & 0.200      & \textbf{0.394}  \\
& BARO       & 0      & 0.208       & 0.208       & 0.124    & 0.083    & 0.133       & 0.171  \\
\thickhline
\end{tabular}
\end{adjustbox}
\label{tab:overall_swat}
\vspace{-5pt}
\end{table}

\begin{table}[!ht]
\renewcommand\arraystretch{1.2}
\small
\caption{Results for RCA baselines on the WADI sub-dataset.} 
\vspace{-2mm}
\centering
\begin{adjustbox}{width=\linewidth}
\begin{tabular}{ccccccccc}
\thickhline
 Dataset & Model  & PR@1 & PR@5 & PR@10 & MRR & MAP@3 & MAP@5 & MAP@10 \\ \thickhline
 
\multirow{7}{*}{WADI} 
& PC & 0.071 & 0.350 & 0.500 & 0.277 & 0.163 & 0.239 & 0.346        \\
 & RCD  & 0 & 0 .071 & 0.119 & 0.054 & 0.048 & 0.057 & 0.076        \\
& $\varepsilon$-Diagnosis   & 0 & 0 & 0.022 & 0.020 & 0 & 0 & 0.009        \\
& CIRCA       & \textbf{0.143}      & \textbf{0.550}   & \textbf{0.714}  & \textbf{0.350}     & \textbf{0.301}    & \textbf{0.400}      & \textbf{0.529}  \\
& BARO      & 0 & 0.143 & 0.143 & 0.085 & 0.071 & 0.100 & 0.121        \\
\thickhline
\end{tabular}
\end{adjustbox}
\label{tab:overall_wadi}
\vspace{-5pt}
\end{table}

\noindent \textbf{Water Treatment/Distribution}.
We evaluate five single-modal RCA methods on the SWaT and WADI sub-datasets using the same set of evaluation metrics. \Cref{tab:overall_swat} presents the results for SWaT, while \Cref{tab:overall_wadi} shows the results for WADI. We observe that CIRCA consistently achieves the best overall performance on both datasets, although the PC algorithm occasionally outperforms it on SWaT by a small margin. This trend aligns with the observations from the Product Review and Cloud Computing sub-datasets. Notably, the results also indicate considerable room for improvement, likely due to the characteristics of the SWaT and WADI datasets—faults are short-lived, and the intervals between them are brief. These fleeting events are easily overlooked by most RCA methods, posing a significant challenge for accurate root cause identification.

\subsection{Online Root Cause Analysis Results}
We evaluate three RCA methods on the Product Review sub-dataset to demonstrate the utility of the \rca in an online setting in Table~\ref{tab:benchmark-online_product_review}.
Due to space limitations, the experimental results on Cloud Computing (Table~\ref{tab:benchmark-online_cloud_computing}) are reported in the Appendix~\ref{more_results}. 
Because the runtime of CIRCA, PC, and Nezha exceed 24 hours per case, they are excluded from the online evaluation. Notably, LEMMA-RCA is a large-scale real-world dataset, consisting of more than 100,000 timestamps across several days with various system fault scenarios, which can be naturally transformed to the online setting, compared with small datasets with limited timestamps for online RCA. Although RCD, $\epsilon$-Diagnosis, and BARO are originally designed for offline analysis, we adapt them to the online setting by formatting the data into a sequence of streaming snapshots. Each baseline method is then evaluated on consecutive snapshots until similar results appear three times in succession or the data stream reaches its final snapshot. The detailed implementation can be found in Appendix~\ref{online_implementation}. Empirically, we observe that all three methods suffer a significant drop in performance under the online setting compared to their offline results. We conjecture that this degradation arises from their inability to consistently capture temporal dependencies across snapshots. This observation highlights the need for developing dedicated online root cause analysis methods. 

\vspace{-2mm}
\begin{table}[htp]
\centering
\small
\renewcommand{\arraystretch}{1.2}
\caption{Results for root cause analysis baselines on Product Review sub-dataset in the online setting.}
\label{tab:benchmark-online_product_review}
\begin{adjustbox}{width=\linewidth}
\begin{tabular}{*{9}{c}}
\thickhline      Modality      & Model         & PR@1      & PR@5      & PR@10     & MRR       & MAP@3     & MAP@5     & MAP@10 \\ \cline{1-9} \hline
\multirow{3}{*}{Metric Only} 
                             & RCD       & 0  &  \textbf{0.250}  &  \textbf{0.250}  &  0.054  &  0  &  0.050  &  0.150  \\
                             & $\epsilon$-Diagnosis       & 0  &  0  &  0  &  0.019  &  0  &  0  &  0  \\
                             & BARO       & \textbf{0.250}  &  \textbf{0.250}  & \textbf{ 0.250}  &  0.269 &  \textbf{0.250}  &  \textbf{0.250}  &  \textbf{0.250}  \\
                             \hline
\multirow{3}{*}{Log Only}    
                             & RCD & 0  &  0  &  0  &  0.012  &  0  &  0  &  0 \\
                             & $\epsilon$-Diagnosis       & 0  & 0  &  0  &  0.025  &  0  &  0.100  &  0.175  \\
                             & BARO        & 0  &  0  &  0  &  0.023  &  0  &  0.100  &  0.175  \\
                             \hline
\multirow{3}{*}{Multi-Modality}    
                                 & RCD       & 0  &  \textbf{0.250}  &  \textbf{0.250}  &  0.054  &  0  &  0.050  &  0.150  \\
                                 & $\epsilon$-Diagnosis       & 0  &  0  &  0  &  0.027  &  0  &  0  &  0  \\
                                 & BARO       & \textbf{0.250}  &  \textbf{0.250}  &  \textbf{0.250}  & \textbf{ 0.270}  &  \textbf{0.250}  &  \textbf{0.250}  &  \textbf{0.250}   \\
% \thickhline
\thickhline
\end{tabular}
\end{adjustbox}
\vspace{-10pt}
\end{table}

\section{Conclusion}
In this work, we introduce LEMMA-RCA, the first large-scale, open-source dataset featuring real system faults across multiple application domains and data modalities. We conduct a comprehensive empirical study using six baseline RCA methods, evaluating their performance on both single-modal and multi-modal data under offline and online settings. The experimental results highlight the utility of LEMMA-RCA as a benchmarking resource. By releasing this dataset publicly, we aim to advance research in root cause analysis for complex systems and support the development of more robust and reliable methodologies, particularly for mission-critical applications.

\bibliographystyle{ACM-Reference-Format}
\bibliography{references}

\appendix

\newpage
\onecolumn
\section{Discussions}
\label{apdx:disc}
\textbf{Broader impact}: To facilitate accurate, efficient, and multi-modal root cause analysis research across diverse domains, we introduce \data\ as a new benchmark dataset. 
Our dataset also offers significant potential for advancing research in areas like \textbf{multi-modal anomaly detection}, \textbf{change point detection}, and \textbf{system diagnosis}. Based on the thorough data analysis and extensive experimental results, we highlight the following areas for future research:
%\vspace{-2mm}
\begin{itemize}[leftmargin=*]

\item \textbf{Expanding Domain Applications}: To enhance the LEMMA-RCA dataset's versatility and impact, we plan to incorporate data from additional domains such as cybersecurity and healthcare. This integration of diverse data sources will facilitate the development of more comprehensive root cause analysis technologies, significantly extending the dataset's applicability across various industries.

\item \textbf{Online and/or Multi-Modal Root Cause Analysis}: Most RCA methods are offline and single-modal, leaving a gap for real-time and/or multi-modal approaches. Developing these methods can enable instant analysis of diverse data streams, essential for dynamic environments like industrial automation and real-time monitoring.

\item \textbf{LLM-Based Root Cause Analysis}: The emergence of LLMs presents new opportunities for RCA by enabling systems to reason over complex, unstructured, and heterogeneous data sources. Future research can explore how LLMs can be adapted or fine-tuned for root cause inference, how they can incorporate domain knowledge, and how their interpretability and reliability can be improved in safety-critical or high-stakes environments.
\end{itemize}

\paragraph{Ethical Considerations.}
LEMMA-RCA does not raise direct ethical concerns, as it is built from system monitoring data rather than human-subject records or personal user data. The dataset is intended to support research on root cause analysis, anomaly detection, and system diagnosis. Its main limitation is coverage: although it includes multiple IT and OT settings, it cannot represent all possible system architectures, workloads, monitoring practices, or failure modes. Therefore, results on LEMMA-RCA should be interpreted within the corresponding domain, modality, and evaluation setting.

%\vspace{-2mm}
\noindent \textbf{Limitations}: Despite its broad capabilities, the LEMMA-RCA dataset may have limited generalizability, as its fault scenarios may not fully capture the diversity of real-world conditions due to factors like system interruptions and unforeseen circumstances. %\textcolor{red} {Additionally, the dependency graphs in our data are semi-complete, reflecting the inherent challenge of obtaining complete ground-truth graphs in complex systems, which may impact the precision of derived analyses.}

\section{More Details of LEMMA-RCA Data} 
\label{dataset_details}
This section outlines the data resources, details the preprocessing steps, and presents visualizations to illustrate the characteristics of the data released. The data licence can be found in~\cref{apx:license}.

\subsection{Data Collection}

We collect real-world data from two domains: IT operations and OT operations. The IT domain includes sub-datasets from Product Review and Cloud Computing microservice systems, while the OT domain includes Secure Water Treatment (SWaT) and Water Distribution (WADI) sub-datasets from water treatment and distribution systems. Data specifics are provided in \Cref{table_aiops_statistics}.

\begin{figure*}
\begin{center}
\begin{tabular}{cc}
\includegraphics[width=0.45\linewidth]{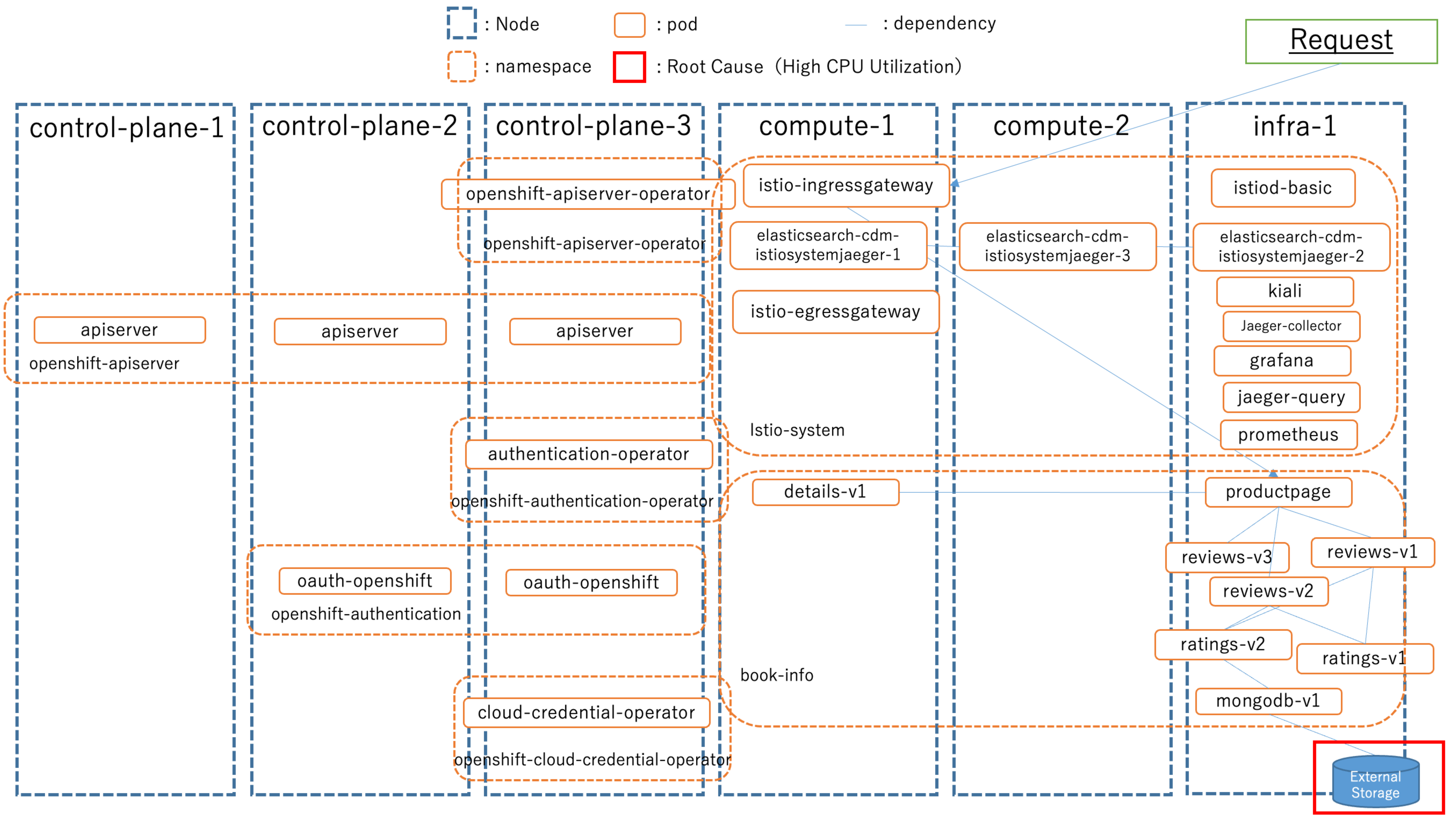} &
\includegraphics[width=0.45\linewidth]{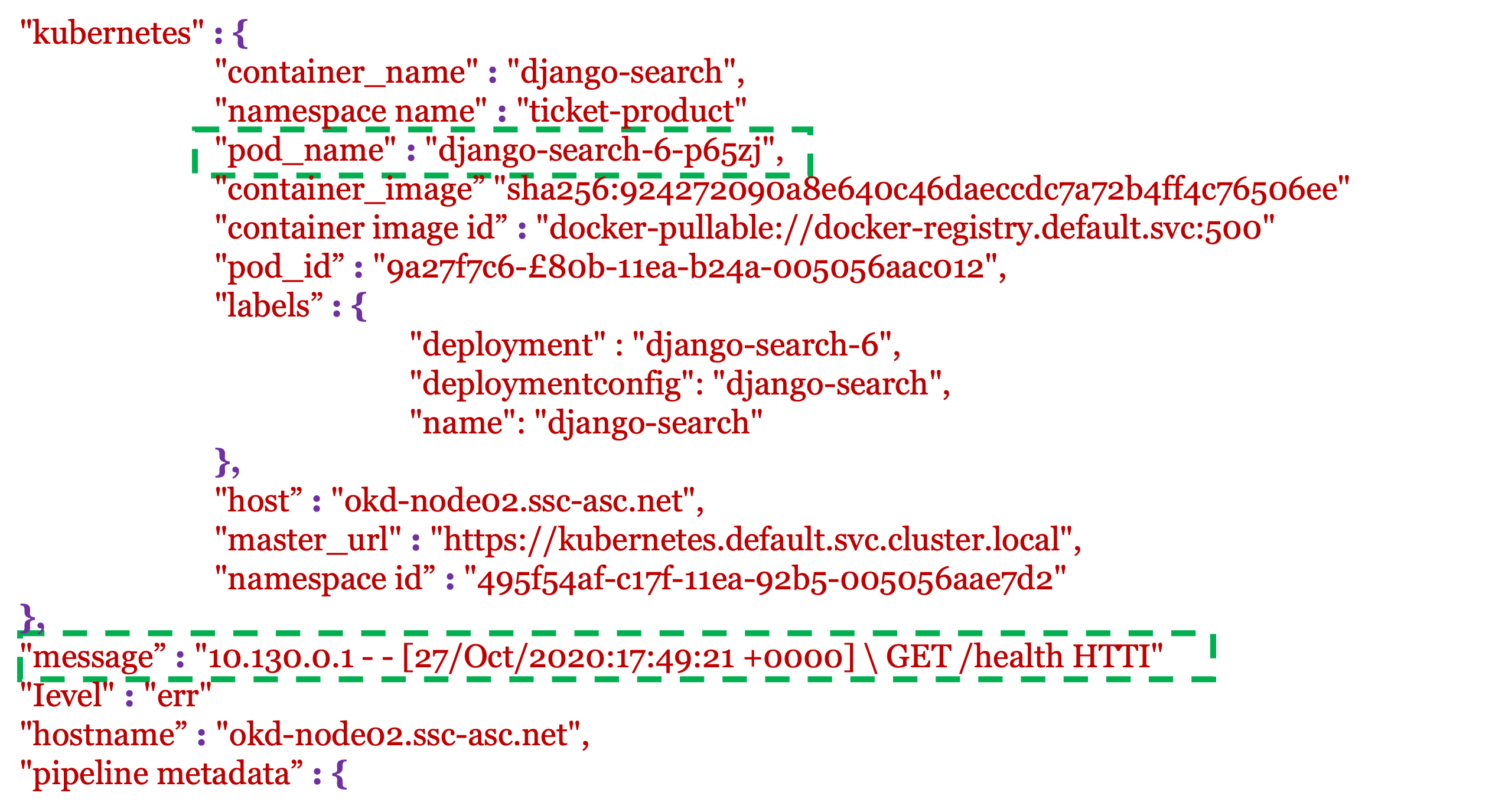} \\
(a) The architecture of Product Review Platform &
(b) Log data captured by the ElasticSearch
\end{tabular}
\end{center}
\vspace{-2mm}
\caption{
Visualization of the microservice system platform, which contains 6 nodes and multiple pods that may vary across different stages; and the ElasticSearch log data. \textbf{We provide the figure with high resolution in the Appendix.}}
\label{fig_aiops_structure}
\end{figure*}

In the IT domain, we developed two microservice platforms: the \textbf{Product Review} Platform and the \textbf{Cloud Computing} Platform. The Product Review Platform is composed of six OpenShift nodes (such as ocp4-control-plane-1 through ocp4-control-plane-3, ocp4-compute-1 and ocp4-compute-2, and ocp4-infra-1) and $216$ system pods (including ProductPage, MongoDB, review, rating, payment, Catalogue, shipping, \textit{etc}.). In this setup, four distinct system faults are collected, including out-of-memory, high-CPU-usage, external-storage-full, and DDoS attack, on four different dates. Each system fault ran the microservice system for at least $49$ hours with different pods involved.
% During the eight days of data collection period, four different types of system faults occurred on separate dates: out-of-memory, high CPU usage, external storage full, and DDoS attacks. 
The pods running in different stages may vary, and the pods associated with different types of system faults also differ.
The structure of this microservice system with some key pods during one fault is depicted in \Cref{fig_aiops_structure} (a). Both log and metric data were generated and stored systematically to ensure comprehensive monitoring. Specifically, eleven types of node-level metrics (\textit{e.g.}, net disk IO usage, net disk space usage, \textit{etc}.) and six types of pod-level metrics (\textit{e.g.}, CPU usage, memory usage, \textit{etc}.) were recorded by Prometheus~\citep{turnbull2018monitoring}, and the time granularity of these system metrics is 1 second. Log data, on the other hand, were collected by ElasticSearch~\citep{zamfir2019systems} and stored in JSON files with detailed timestamps and retrieval periods. The contents of system logs include timestamp, pod name, log message, \textit{etc}., as shown in \Cref{fig_aiops_structure} (b). The JMeter~\citep{nevedrov2006using} was employed to collect the system status information, such as elapsed time, latency, connect time, thread name, throughput, \textit{etc}. The latency is considered as system KPI as the system failure would result in the latency significantly increasing.

For the Cloud Computing Platform, we monitored six different types of faults (such as cryptojacking, mistakes made by GitOps, configuration change failure, \textit{etc}.), and collected system metrics and logs from various sources. In contrast to the Product Review platform, system metrics were directly extracted from CloudWatch\footnote{https://aws.amazon.com/cloudwatch/} Metrics on EC2 instances, and the time granularity of these system metrics is 1 second. Log events were acquired from CloudWatch Logs, consisting of three data types (\textit{i.e.}, log messages, api debug log, and mysql log). Log message describes general log message about all system entities; api debug log contains debug information of the AP layer when the API was executed; mysql logs contain log information from database layer, including connection logs to mysql, which user connected from which host, and what queries were executed. Latency, error rate, and utilization rate were tracked using JMeter tool, serving as Key performance indicators (KPIs). This comprehensive logging and data storage setup facilitates detailed monitoring and analysis of the system's performance and behavior. 

% \begin{table}[!ht]
% \vspace{-15pt}
% \renewcommand\arraystretch{1.1}
% \scriptsize
% \centering
% \caption{Data statistics of IT and OT operation sub-datasets.} 
% \vspace{-2mm}
% \begin{adjustbox}{width=\linewidth}
% \begin{tabular}{ccc}
% \thickhline
% Microservice System (IT) &  Product Review  & Cloud Computing \\ \hline
% Original Dataset Size & 765 GB & 540 GB  \\ 
% Number of (\#) fault types &  4 & 6 \\ 
% %Mean of Number of nodes &  6 & 9.67 \\ \hline
% Average \# entities per fault &  216.0 & 167.71 \\ 
% Average \# metrics per fault  & 11 (node-level) + 6 (pod-level) & 6 (node-level) + 7 (pod-level) \\ 
% Average  \# timestamps per fault & 131,329.25 & 109,350.57 \\  
% Average max log events per fault across pods& 153,081,219.0 & 63,768,587.25\\ 
% \thickhline
% Water Treatment/Distribution (OT) &  SWaT  & WADI \\ \hline
% Original Dataset Size & 4.47G & 5.67G  \\ 
% Number of (\#)fault types &  16 & 9 \\ 
% Average \# entities per fault &  51.0 & 123.0 \\ 
% Average \# metrics per fault  & 7 (node-level) + 7 (pod-level) & 7 (node-level) + 7 (pod-level) \\ 
% Average \# timestamps per fault&  56239.88 & 85248.47 \\ 
% \thickhline
% \end{tabular}
% \end{adjustbox}
% \label{table_aiops_statistics}
% \end{table}

In the OT domain, we constructed two sub-datasets, SWaT and WADI, using monitoring data collected by the iTrust lab at the Singapore University of Technology and Design~\citep{swat_wadi}. These two sub-datasets consist of time-series/metrics data, capturing the monitoring status of each sensor/actuator as well as the overall system at each second. Specifically, SWaT~\citep{mathur2016swat} was collected over an 11-day period from a water treatment testbed equipped with $51$ sensors. The system operated normally during the first $7$ days, followed by attacks over the last $4$ days, resulting in 16 system faults. Similarly, WADI~\citep{ahmed2017wadi} was gathered from a water distribution testbed over $16$ days, featuring $123$ sensors and actuators. The system maintained normal operations for the first $14$ days before experiencing attacks in the final $2$ days, with $15$ system faults recorded.

% \begin{figure*}[h]
% \begin{center}
% \includegraphics[width=0.9\linewidth]
% {Figure/kpi_visualization.jpg}
% \end{center}
% \vspace{-1mm}
% \caption{
% Visualization of KPI for system failure cases. \textbf{Left}: the first two sub-figures are from the Product Review sub-dataset; the third and fourth sub-figures are from the Cloud Computing sub-dataset; \textbf{Right}: the first two sub-figures are from the SWaT sub-dataset; the last two sub-figures are from the WADI sub-dataset.}
% \label{fig_aiops}
% \end{figure*}

We visualized the key performance indicator (KPI) for eight failure cases in \Cref{fig_aiops}, where sudden spikes or drops in latency indicate system failures. The first two sub-figures on the left show the KPIs for two faults in the Product Review sub-dataset, while the third and fourth sub-figures depict faults in the Cloud Computing sub-dataset. The first two sub-figures on the right display faults in the SWaT dataset, and the last two show faults in the WADI dataset. The x-axis represents the timestamp, and the y-axis shows the system latency.

\subsection{Data Preprocessing}
After collecting system metrics and logs, we assess whether each pod exhibits stationarity, as non-stationary data are unpredictable and cannot be effectively modeled. Consequently, we exclude non-stationary pods, retaining only stationary ones for subsequent data preprocessing steps.

\textbf{Log Feature Extraction for Product Review and Cloud Computing.} The logs of some system entities we collected are limited and insufficient for meaningful root cause analysis. Thus, we exclude them from further analysis. Additionally, the log data is unstructured and frequently uses a special token, complicating its direct application for analysis. How to extract useful information from unstructured log data remains a great challenge. Following~\citep{zheng2025online}, we preprocess the log data into time-series format. We first utilize a log parsing tool, such as Drain, to transform unstructured logs into structured log messages represented as templates. We then segment the data using fixed 10-minute windows with 30-second intervals, calculating the occurrence frequency of each log template. This frequency forms our first feature type, denoted as $X^L_1\in \mathbb{R}^{T}$, where $T$ is the number of timestamps. We prioritize this feature because frequent log templates often indicate critical insights, particularly useful in identifying anomalies such as Distributed Denial of Service (DDoS) attacks, where a surge in template frequency can indicate unusual activity.

Moreover, we introduce a second feature type based on `golden signals' derived from domain knowledge, emphasizing the frequency of abnormal logs associated with system failures like DDoS attacks, storage failures, and resource over-utilization. Identifying specific keywords like `error,' `exception,' and `critical' within log templates helps pinpoint anomalies. This feature, denoted as $X^L_2\in \mathbb{R}^{T}$, assesses the presence of abnormal log templates to provide essential labeling information for anomaly detection.

Lastly, we implement a TF-IDF based method, segmenting logs using the same time windows and applying Principal Component Analysis (PCA) to reduce feature dimensionality, selecting the most significant component as $X^L_3\in \mathbb{R}^{T}$. We concatenate these three feature types to form the final feature matrix $X^L = [X^L_1; X^L_2; X^L_3] \in \mathbb{R}^{3 \times T}$, enhancing our capacity for a comprehensive analysis of system logs and improving anomaly detection capabilities.

\textbf{KPI Construction for SWaT and WADI}. The SWaT and WADI sub-datasets include the label column that reflects the system status; however, the values within this column are discrete. To facilitate the root cause analysis, it is beneficial to transform these values into a continuous format. Specifically, we propose to convert the label into a continuous time series. To achieve this, we employ anomaly detection algorithms, such as Support Vector Data Description and Isolation Forest, to model the data. Subsequently, the anomaly score, as determined by the model, will be utilized as the system KPI. More data preprocessing details on SWaT and WADI can be found in~\Cref{ape:time}.

\subsection{System Fault Scenarios}
\label{system_faul_cases}
\begin{figure*}[h!]
\begin{center}
\begin{tabular}{cc}
\includegraphics[width=0.45\linewidth]{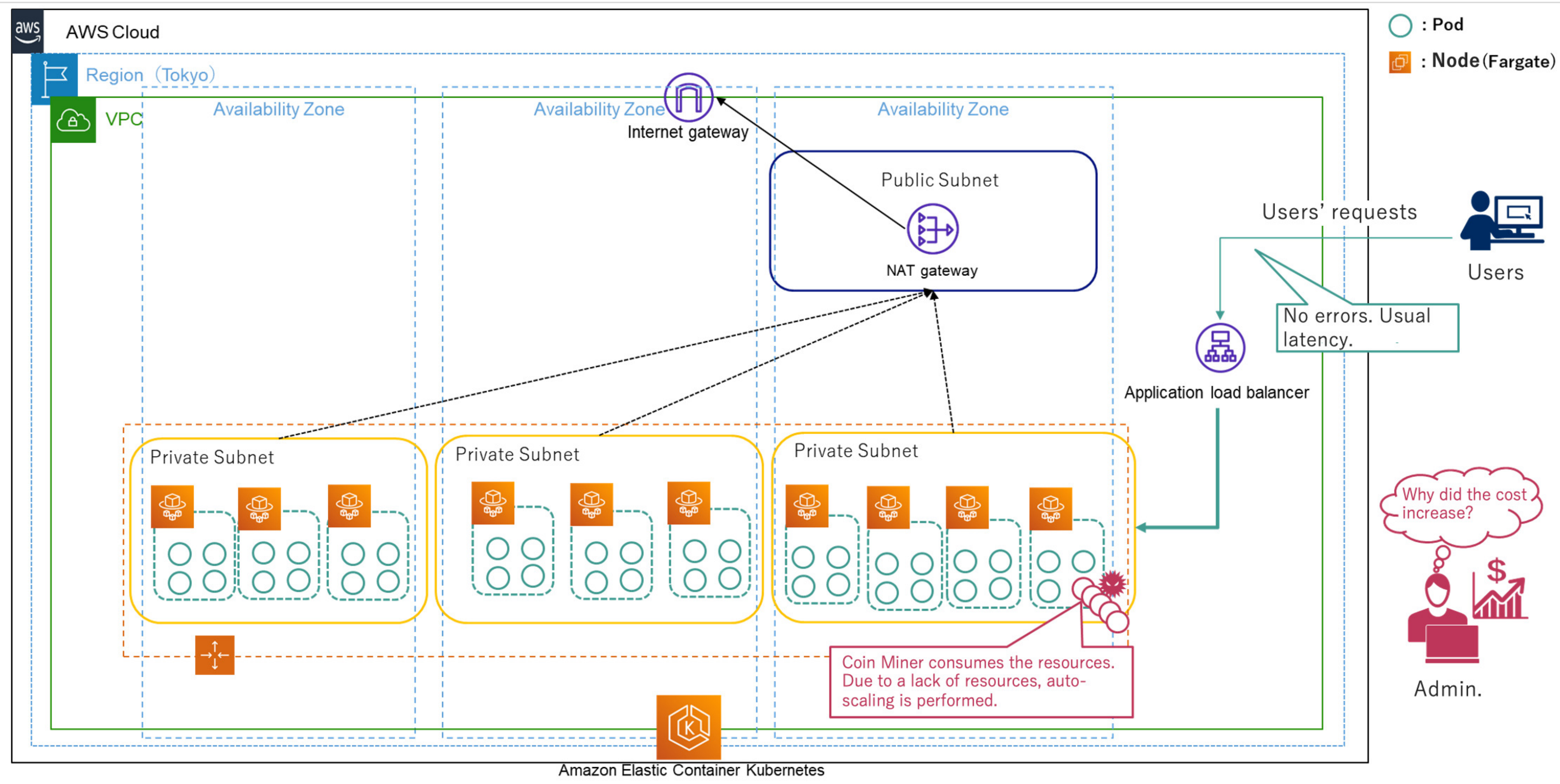} &
\includegraphics[width=0.45\linewidth]{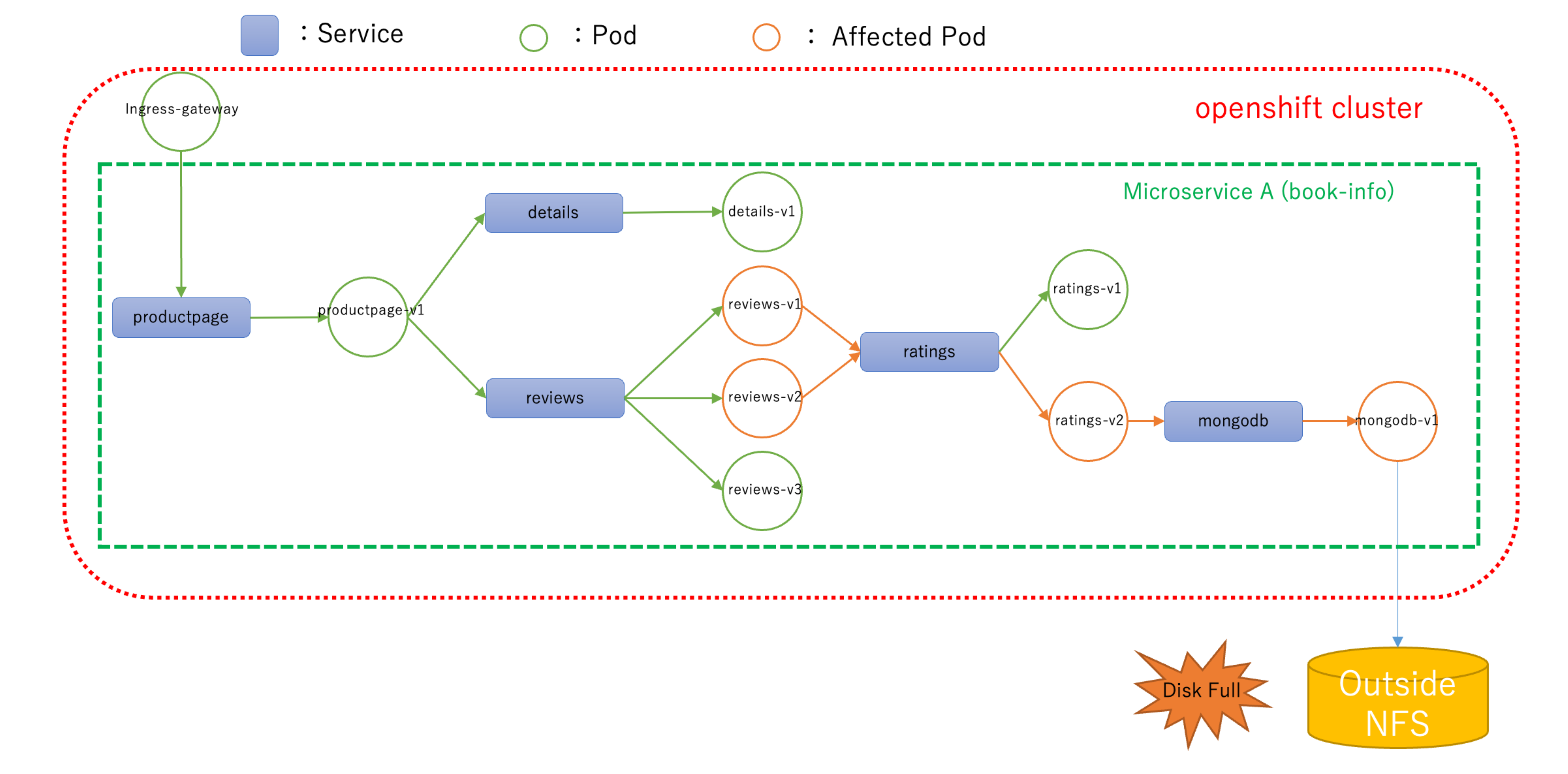} \\
\end{tabular}
\end{center}
\vspace{-2mm}
\caption{
Visualization of two system fault scenarios. \textbf{Left:} Cryptojacking. \textbf{Right:} External storage failure. }
\label{fig_aiops_scenarios}
%\vspace{-10pt}
\end{figure*}

There are $10$ different types of real system faults in Product Review and Cloud Computing sub-datasets. We also visualize the system fault of these two cases in \Cref{fig_aiops_scenarios}.
%\vspace{-2mm}

\begin{figure*}[h]
%\vspace{-25pt}
\begin{center}
\begin{tabular}{cc}
\includegraphics[width=0.49\linewidth]
{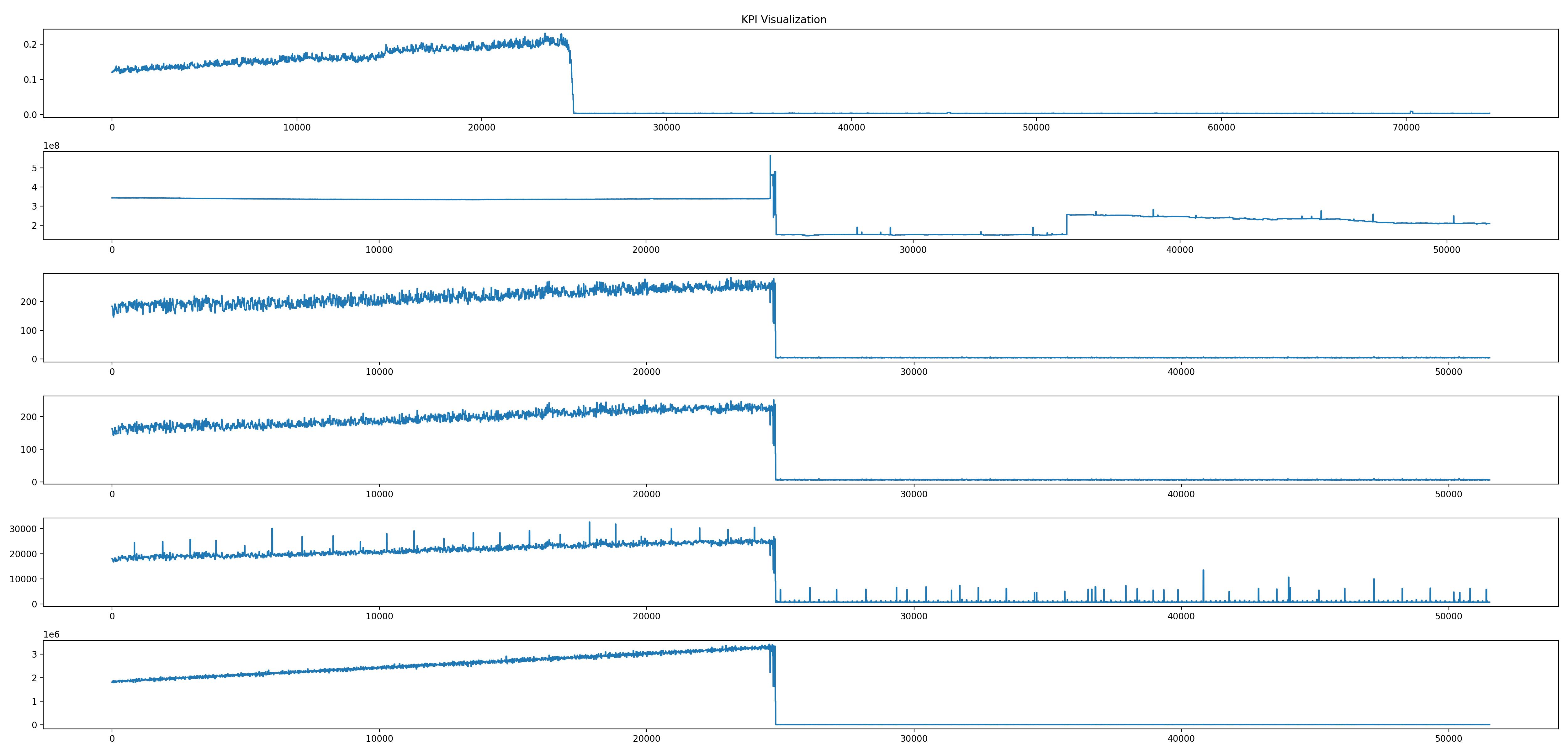} &  \includegraphics[width=0.49\linewidth]
{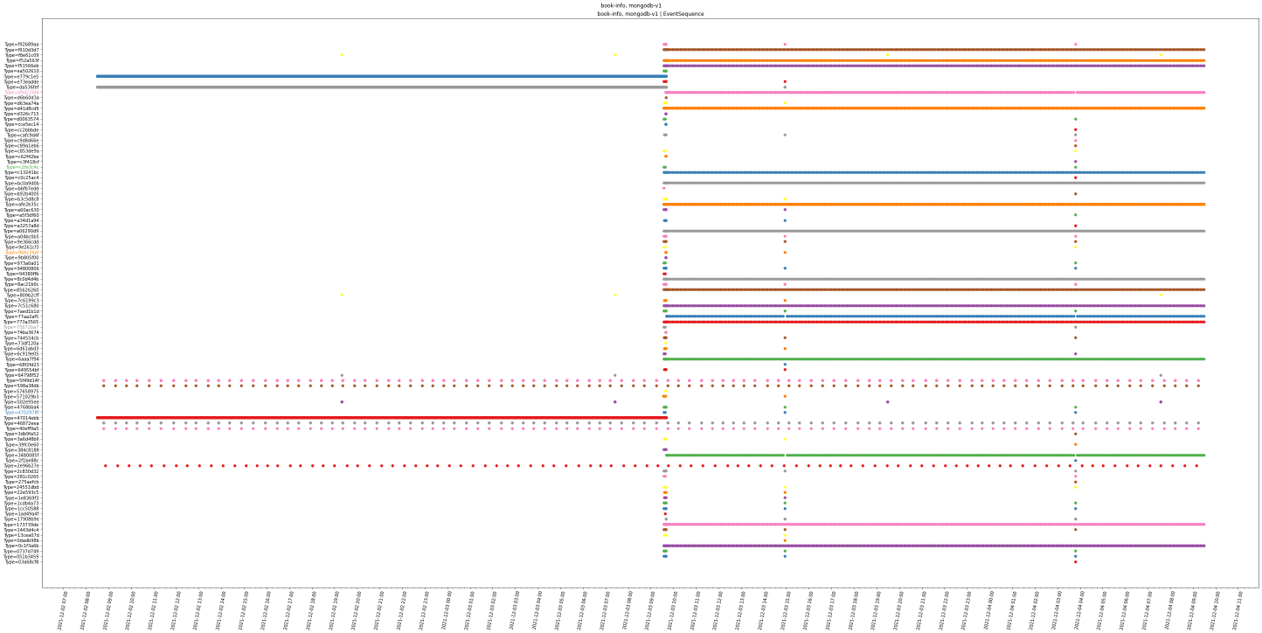}\\
\end{tabular}
\end{center}
%\vspace{-2mm}
\caption{
Visualization of root cause for one system failure case (\textit{i.e.}, \textbf{External Storage Failure}) on the Product Review Platform. \textbf{Left:} six system metrics of root cause. \textbf{Right:} the system log of the root cause pod (\textit{i.e.}, Mongodb-v1) with the x-axis representing the timestamp, the y-axis indicating the log event ID, and the colored dots denoting event occurrences. Sudden drops in the metrics data, as well as new log event patterns observed at the midpoint, indicate a system failure.}
\label{fig_1203_visualization}
\end{figure*}

%\vspace{-2mm}
\begin{itemize}[leftmargin=*]
\item \textbf{Cryptojacking}. 
    \begin{itemize}
    \item \textbf{Description:} 
        In this scenario, cloud usage fees increase due to cryptojacking, where a Coin Miner is covertly downloaded and installed on a microservice (details-v1 pod) in an EKS cluster. This miner gradually consumes IT resources, escalating the cloud computing costs. Identifying the root cause is challenging because the cost (SLI) encompasses the entire system, and no individual service errors are detected. 
        \item \textbf{Method:} We periodically send the external requests to microservices, and after a day, the miner's activity triggers auto-scaling in details-v1, increasing resource usage. Fargate's impact on EKS costs is significant due to its resource dependency. 
        \item \textbf{Data Collection:} KPI (SLI) is calculated from resource usage, with all pod and node metrics collected from CloudWatch. However, there are no node logs for Fargate, complicating diagnosis.
    \end{itemize}

    \item\textbf{External Storage Failure}. 
    \begin{itemize}
        \item \textbf{Description:} The external storage disk becomes full, causing the failure of adding new data by the Database (DB) pod (\textit{i.e.}, mongodb-v1) and resulting in sytem errors. These errors propagate to pods that depend on the DB pod, causing some services (ratings) within Microservice A to encounter errors. 
        \item \textbf{Method:} We fill up the external storage disk connected to the Database (DB) pod (\textit{i.e.}, mongodb-v1) within Microservice A's OpenShift\footnote{https://www.redhat.com/en/technologies/cloud-computing/openshift} cluster. 
        \item \textbf{Data Collection:} We monitor changes in response and error information for Microservice A using Jaeger logs. Metrics for all containers and nodes, including CPU and memory usage, are obtained from Prometheus within OpenShift. Logs for all containers and nodes are retrieved from Elasticsearch within OpenShift. Additionally, we collect message logs from the external storage. We illustrate the metrics and log data of the root cause pod in~\Cref{fig_1203_visualization}.
    \end{itemize}

% This section describes the processes used to generate and monitor system fault scenarios, with emphasis on mimicking real-world fault patterns. Each scenario involved the induction of specific failure conditions, while allowing the microservice system to exhibit its natural behavior under stress. Metrics and logs were collected using established monitoring tools, such as Prometheus, Elasticsearch, CloudWatch, Jaeger, and JMeter.

% \begin{itemize}
    \item \textbf{Silent Pod Degradation Fault.}  
    \begin{itemize}
        \item \textbf{Description:} A pod in a load balancer contains a latent bug causing its CPU usage to rise, which gradually increases latency for a subset of users without triggering autoscaling or error alerts.  
        \item \textbf{Method:} We periodically sent requests to Microservice A over a 24-hour period. After this initial observation, we manually increased the CPU load on one specific \texttt{productpage-v1} pod to simulate the bug.  
        \item \textbf{Data Collection:} Metrics and logs were collected from CloudWatch, while KPIs such as latency were measured using JMeter. The goal was to trace latency increases back to the specific pod with elevated CPU utilization.  
    \end{itemize}

    \item \textbf{Noisy Neighbor Issue.}  
    \begin{itemize}
        \item \textbf{Description:} A neighboring pod in a shared node generates high CPU load, impacting the performance of the \texttt{productpage-v1} pod and causing elevated error rates.  
        \item \textbf{Method:} Requests were sent to Microservice A, while the pod ratings of Microservice B (\texttt{robot-shop}) were moved to the same node as \texttt{productpage-v1}, generating contention.  
        \item \textbf{Data Collection:} Metrics (CPU usage, memory usage) were gathered using Prometheus, while logs were obtained from CloudWatch Logs. Configuration changes, such as node assignments, were also recorded.  
    \end{itemize}

    \item \textbf{Node Resource Contention Stress Test.}  
    \begin{itemize}
        \item \textbf{Description:} A stress test on CPU resources was conducted by inducing high load on Microservice B, co-located with Microservice A on the same node.  
        \item \textbf{Method:} Periodic requests were sent to Microservice A using JMeter, while a high CPU load was generated on Microservice B using the \texttt{OpenSSL speed} command.  
        \item \textbf{Data Collection:} HTTP response logs from JMeter were analyzed for performance impacts. System metrics (CPU and memory usage) were retrieved from Prometheus, while container logs were collected from Elasticsearch.  
    \end{itemize}

    \item \textbf{DDoS Attack.}  
    \begin{itemize}
        \item \textbf{Description:} A Distributed Denial of Service (DDoS) attack was simulated to overload the system, causing Out-of-Memory (OOM) errors in targeted pods.  
        \item \textbf{Method:} Over a monitoring period of approximately 48 hours, we gradually increased the request rate to Microservice A, eventually overwhelming the \texttt{reviews-v2} and \texttt{reviews-v3} pods.  
        \item \textbf{Data Collection:} Metrics such as CPU and memory utilization were collected via Prometheus. Logs from Jaeger and Elasticsearch provided insights into the system’s response to the attack.  
    \end{itemize}

    \item \textbf{Malware Attack.}  
    \begin{itemize}
        \item \textbf{Description:} A malware pod executed a password list attack to compromise other pods, propagating DDoS scripts to degrade overall system performance.  
        \item \textbf{Method:} The attack started from a designated pod (i.e., \texttt{scenario10-malware-deployment}) and targeted others via SSH password brute-forcing, ultimately generating high load on \texttt{productpage-v1}.  
        \item \textbf{Data Collection:} JMeter was used to monitor KPIs (latency, error rate), while Prometheus and CloudWatch Logs provided system metrics and logs for root-cause analysis.  
    \end{itemize}

    \item \textbf{Bug Infection.}  
    \begin{itemize}
        \item \textbf{Description:} A latent bug in the API caused asymmetric CPU load increases, degrading response times without fully utilizing the CPU capacity.  
        \item \textbf{Method:} Requests were sent periodically to the web service, and after a day, a script induced increased CPU utilization on one core.  
        \item \textbf{Data Collection:} KPIs were measured using JMeter, while system metrics and logs were collected via CloudWatch for detailed analysis.  
    \end{itemize}

    \item \textbf{Configuration Fault.}  
    \begin{itemize}
        \item \textbf{Description:} An incorrect resource limit in a Kubernetes manifest file led to a pod being terminated by the OOM killer, impacting other services.  
        \item \textbf{Method:} Requests were sent to Microservice A, while a Git push introduced a faulty configuration for the \texttt{details-v1} pod. The misconfigured pod eventually failed under load.  
        \item \textbf{Data Collection:} Error rates were tracked using JMeter, and metrics/logs were retrieved from Prometheus and CloudWatch for root-cause identification.  
    \end{itemize}
    
\end{itemize}

\section{Implementation Details of Online Setting}
\label{online_implementation}
In this section, we describe how we format the data for the online setting and adapt offline RCA methods accordingly. For each case, the data is split chronologically into two parts: historical data and streaming data. The split point is determined by the “online start,” a timestamp prior to the system fault (typically 1–2 hours before the failure), indicating that a fault is imminent. The streaming data is further divided into $K$ snapshots, each covering a fixed time window of 400 seconds. To construct the first batch, we combine the historical data with the first snapshot, which serves as input for each baseline method to identify the root cause. We then update the batch iteratively by sliding the window: at each step, we remove the earliest 400 seconds from the historical data and replace it with the next 400 seconds from the streaming snapshots, ensuring the total batch length remains fixed. Offline methods adapted to this setting are evaluated sequentially on the snapshots until either (i) consistent results are obtained for three consecutive steps, or (ii) the data stream reaches its final snapshot.

\section{Monitoring Time Series Segmentation for SWaT and WADI} 
\label{ape:time}

In the original SWaT and WADI datasets, the attack model demonstrates irregular attack patterns, occasionally targeting multiple sensors simultaneously, or executing attacks at closely spaced intervals. To follow the principles of RCA, we have established two specific preprocessing rules for these datasets:
1) Each recorded attack event must only involve a single sensor or actuator.
2) The duration of the dataset corresponding to each attack event must be standardized to two hours.
Consequently, we selectively keep attack events that impact only one sensor or actuator. If the interval between successive attack events is insufficiently short, we assume the stability in the monitoring data immediately before and after each attack event. To ensure the necessary two-hour duration for each event, we concatenate normal-state data from both before and after the attack period. This adjustment positions the attack event centrally within a continuous two-hour segment, facilitating consistent and accurate analysis. 

\section{More Experimental Results}
\label{more_results}
We present the results for RCA on Cloud Computing sub-dataset in the offline setting in Table~\ref{table_result_2} and in the online setting in Table~\ref{tab:benchmark-online_cloud_computing}.

\begin{table}[ht]
\renewcommand{\arraystretch}{1.2}
\caption{Results for RCA with multiple modalities on the Cloud Computing sub-dataset.}
%\vspace{-2mm}
\small
\centering
\begin{adjustbox}{width=0.75\linewidth}
\begin{tabular}{*{9}{c}}
\thickhline      Modality      & Model         & PR@1      & PR@5      & PR@10     & MRR       & MAP@3     & MAP@5     & MAP@10 \\ \cline{1-9} \hline
\multirow{5}{*}{Metric Only} 
% & Dynotears & 0 & 0.167 & 0.333 & 0.075 & 0 & 0.033 & 0.117 \\
                             & PC & 0 & 0 & 0 & 0.029 & 0 & 0 & 0 \\
                             % & C-LSTM & 0.167 & 0.333 & 0.333 & 0.300 & 0.278 & 0.300 & 0.317 \\
                             % & GOLEM & 0 & 0 & 0.167 & 0.044 & 0 & 0 & 0.017 \\
                             & RCD       & 0  &  \textbf{0.250}  &  0.250  &  0.126  &  \textbf{0.083}  &  \textbf{0.150}  &  0.200  \\
                             & $\epsilon$-Diagnosis       & 0  &  0  &  0.250  &  0.067  &  0  &  0  &  0.025  \\
                             & CIRCA       & 0  &  0  &  \textbf{0.500}  &  0.105  &  0  &  0  &  0.150  \\
                             & BARO       & 0  &  \textbf{0.250}  &  0.250  &  0.105  &  0  &  0.100  &  0.175  \\
                             % & REASON & 0.167 & \textbf{1.000} & \textbf{1.000} & 0.472 & 0.444 & 0.667 & 0.833 \\
                             \hline
\multirow{5}{*}{Log Only}    
                             & PC & 0 & 0 & 0 & 0.032 & 0 & 0 & 0 \\
                             & RCD & 0  &  0  &  0  &  0.059  &  0  &  0  &  0 \\
                             & $\epsilon$-Diagnosis & 0  &  0  &  0  &  0.059  &  0  &  0  &  0 \\
                             & CIRCA       & 0  &  \textbf{0.250}  &  0.250  &  0.110  &  0  &  0.100  &  0.175  \\
                             & BARO        & 0  &  \textbf{0.250}  &  0.250  &  0.105  &  0  &  0.100  &  0.175  \\
                             % & C-LSTM & 0 & 0 & 0.167 & 0.044 & 0 & 0 & 0.050 \\
                             % & GOLEM & 0 & 0 & 0.167 & 0.051 & 0 & 0 & 0.050 \\
                             % & REASON & 0 & 0 & 0.333 & 0.082 & 0 & 0 & 0.067 \\
                             \hline
% \multirow{7}{*}{Multi-Modality} & Dynotears & 0 & 0.167 & 0.333 & 0.095 & 0 & 0.033 & 0.015 \\
%                              & PC & 0 & 0 & 0.167 & 0.042 & 0 & 0 & 0.050 \\
%                              & C-LSTM & 0.167 & 0.333 & 0.500 & 0.267 & 0.167 & 0.233 & 0.367 \\
%                              & GOLEM & 0 & 0 & 0.333 & 0.075 & 0 & 0 & 0.083 \\
%                              & REASON & 0.333 & \textbf{1.000} & \textbf{1.000} & 0.597 & 0.611 & 0.767 & 0.883 \\
%                              & Nezha & 0 & 0.333 & 0.333 & 0.148 & 0.111 & 0.020 & 0.267 \\
\multirow{7}{*}{Multi-Modality}  & PC                 & 0     & 0     & 0.250  & 0.064   & 0        & 0     & 0.125     \\    
                                 & RCD	             & 0  &  \textbf{0.250}  &  0.250  &  0.092  &  0  &  0.050  &  0.150  \\ 
                                 & $\epsilon$-Diagnosis	 & 0  &  0  &  0.250  &  0.067  &  0  &  0  &  0.025     \\ 
                                 & CIRCA	             & 0  &  \textbf{0.250}  &  \textbf{0.500}  &  \textbf{0.147}  &  \textbf{0.083}  &  \textbf{0.150}  &  \textbf{0.225}  \\ 
                                 & BARO	             & 0  &  \textbf{0.250}  &  0.250  &  0.126  & \textbf{ 0.083}  &  \textbf{0.150}  &  0.200   \\
                                 & Nezha	             & 0	 & \textbf{0.250}	 & 0.250	 & 0.105	 & 0	    & 0.100	 & 0.175 \\ 
\thickhline
\end{tabular}
\end{adjustbox}
\label{table_result_2}
\end{table}

%\vspace{-2mm}
\begin{table}[h!]
\centering
\small
\renewcommand{\arraystretch}{1.2}
\caption{Results for root cause analysis baselines on Cloud Computing sub-dataset in the online setting.}
\label{tab:benchmark-online_cloud_computing}
\begin{adjustbox}{width=0.75\linewidth}
\begin{tabular}{*{9}{c}}
\thickhline      Modality      & Model         & PR@1      & PR@5      & PR@10     & MRR       & MAP@3     & MAP@5     & MAP@10 \\ \cline{1-9} \hline
\multirow{3}{*}{Metric Only} 
                             & BARO                             & 0  &  \textbf{0.250}  &  \textbf{0.500}  &  0.172 &  \textbf{0.167} &  \textbf{0.200}  &  \textbf{0.275}  \\
                             & $\epsilon$-Diagnosis             & 0  &  \textbf{0.250}  &  \textbf{0.500}  &  0.173  &  \textbf{0.167}  &  \textbf{0.200}  &  0.250  \\
                             & RCD                              & 0  &  \textbf{0.250}  &  0.250  &  0.162  &  \textbf{0.167}  &  \textbf{0.200}  &  0.225  \\
                             \hline
\multirow{3}{*}{Log Only}    
                             & BARO                             & 0  &  0  &  0.250  &  0.067  &  0  &  0  &  0.075 \\
                             & $\epsilon$-Diagnosis             & 0  &  0  &  0.250  &  0.072  &  0  &  0  &  0.100  \\
                             & RCD                              & 0  &  0  &  0  &  0.044  &  0  &  0  &  0 \\
                             \hline
\multirow{3}{*}{Multi-Modality}    
                                 & BARO                         & 0  &  \textbf{0.250}  &  \textbf{0.500}  &  0.173  &  \textbf{0.167}  &  \textbf{0.200}  &  \textbf{0.275}   \\
                                 & $\epsilon$-Diagnosis         & 0  &  \textbf{0.250}  &  \textbf{0.500}  &  \textbf{0.181}  &  \textbf{0.167}  &  \textbf{0.200}  &  0.250  \\
                                 & RCD                          & 0  &  \textbf{0.250}  &  0.250  &  0.162  &  \textbf{0.167}  &  \textbf{0.200}  &  0.225  \\
% \thickhline
\thickhline
\end{tabular}
\end{adjustbox}
%\vspace{-10pt}
\end{table}

\section{LEMMA-RCA License}
\label{apx:license}
%We release our dataset under the CC BY-NC 4.0 International License License, hoping to contribute to the broader research community.
The \rca benchmark dataset is released under a CC BY-ND 4.0 International License: https://creativecommons.org/licenses/by-nd/4.0. %Our data preprocessing code is released under theMIT License: https://opensource.org/licenses/MIT. 
The license of any specific baseline methods used in our codebase should be verified on their official repositories.

\section{Reproducibility} 
\label{ape:rep}
All experiments are conducted on a server running Ubuntu 18 with an Intel(R) Xeon(R) Silver 4110 CPU @2.10GHz and one 11GB GTX2080 GPU. %In the online RCA experiment, we set the size of historical metric and log data to 8-hour intervals and each batch is set to be a $10$-minute interval. We use the Adam as the optimizer and we train the model for 100 iterations at each batch. 
In addition, all methods were implemented using Python 3.8.12 and PyTorch 1.7.1.

\section{Detailed Description of Baselines}
\label{apx:baseline}

We evaluate the performance of the following RCA models on the benchmark sub-datasets: 
\begin{itemize}
    %\item \textbf{PC}~\citep{DBLP:journals/technometrics/Burr03}: The PC algorithm is a data-driven method for causal discovery, producing a partially directed acyclic graph (PDAG) that represents causal relationships among variables. It starts with a fully connected graph and iteratively removes edges based on conditional independence tests, then orients the remaining edges to construct a causal structure. The algorithm assumes the causal Markov property, no hidden confounders, and no cycles in the graph. It is widely used for root cause analysis to identify direct and indirect influences on specific outcomes but is sensitive to the reliability of independence tests and cannot distinguish between equivalent causal structures.
    \item \textbf{PC-based}~\citep{ma2020automap}: It is an unsupervised root cause analysis method designed for complex systems using multiple types of performance metrics. It constructs an anomaly behavior graph using a PC-based causal discovery algorithm to capture service-level dependencies. Based on this graph, it applies heuristic random walk strategies—including forward, backward, and self-directed walks—to trace and rank potential root causes. 

    \item \textbf{CIRCA}~\citep{li2022causal}: CIRCA is an unsupervised root cause analysis method that formulates the problem as a causal inference task called intervention recognition. Its core idea is to identify root cause indicators by evaluating changes in the probability distribution of monitoring variables conditioned on their parents in a Causal Bayesian Network (CBN). CIRCA applies this approach to online service systems by constructing a graph among monitoring metrics, leveraging system architecture knowledge and causal assumptions to guide the analysis.
    \item \textbf{$\epsilon$-Diagnosis}~\citep{shan2019diagnosis}: $\epsilon$-Diagnosis is an unsupervised, low-cost diagnosis algorithm designed to address small-window long-tail latency (SWLT) in web services, which arises in short statistical windows and typically affects a small subset of containers in microservice clusters. It uses a two-sample test algorithm and $\epsilon$-statistics to measure the similarity of time series, enabling the identification of root-cause metrics from millions of metrics. The algorithm is implemented in a real-time diagnosis system for production microservice platforms.
    \item \textbf{RCD}~\citep{ikram2022root}: RCD is a scalable algorithm for detecting root causes of failures in complex microservice architectures using a hierarchical and localized learning approach. It treats the failure as an intervention to quickly identify the root cause, focuses learning on the relevant portion of the causal graph to avoid costly conditional independence tests, and explores the graph hierarchically. The technique is highly scalable, providing actionable insights about root causes, while traditional methods become infeasible due to high computation time.
    \item \textbf{BARO}~\citep{pham2024baro}: BARO is a robust, end-to-end RCA framework designed for multivariate time-series data in microservice systems. It integrates anomaly detection and root cause localization using Multivariate Bayesian Online Change Point Detection (BOCPD) to detect failures and estimate their occurrence times. For RCA, BARO introduces RobustScorer, a nonparametric hypothesis testing approach that ranks candidate root causes based on their distributional shifts, using median and interquartile range rather than mean and standard deviation to improve robustness.

       \item \textbf{Nezha}~\citep{yu2023nezha}: Nezha is an interpretable and fine-grained root cause analysis (RCA) method for microservices that unifies heterogeneous observability data (metrics, traces, logs) into a homogeneous event format. This representation enables the construction of event graphs for integrated analysis. Nezha statistically localizes actionable root causes at granular levels, such as specific code regions or resource types, offering high interpretability to support confident mitigation actions by SREs.
\end{itemize}

\section{Figures for Clarity}
\begin{figure}
    \centering
    \includegraphics[width=1\linewidth]{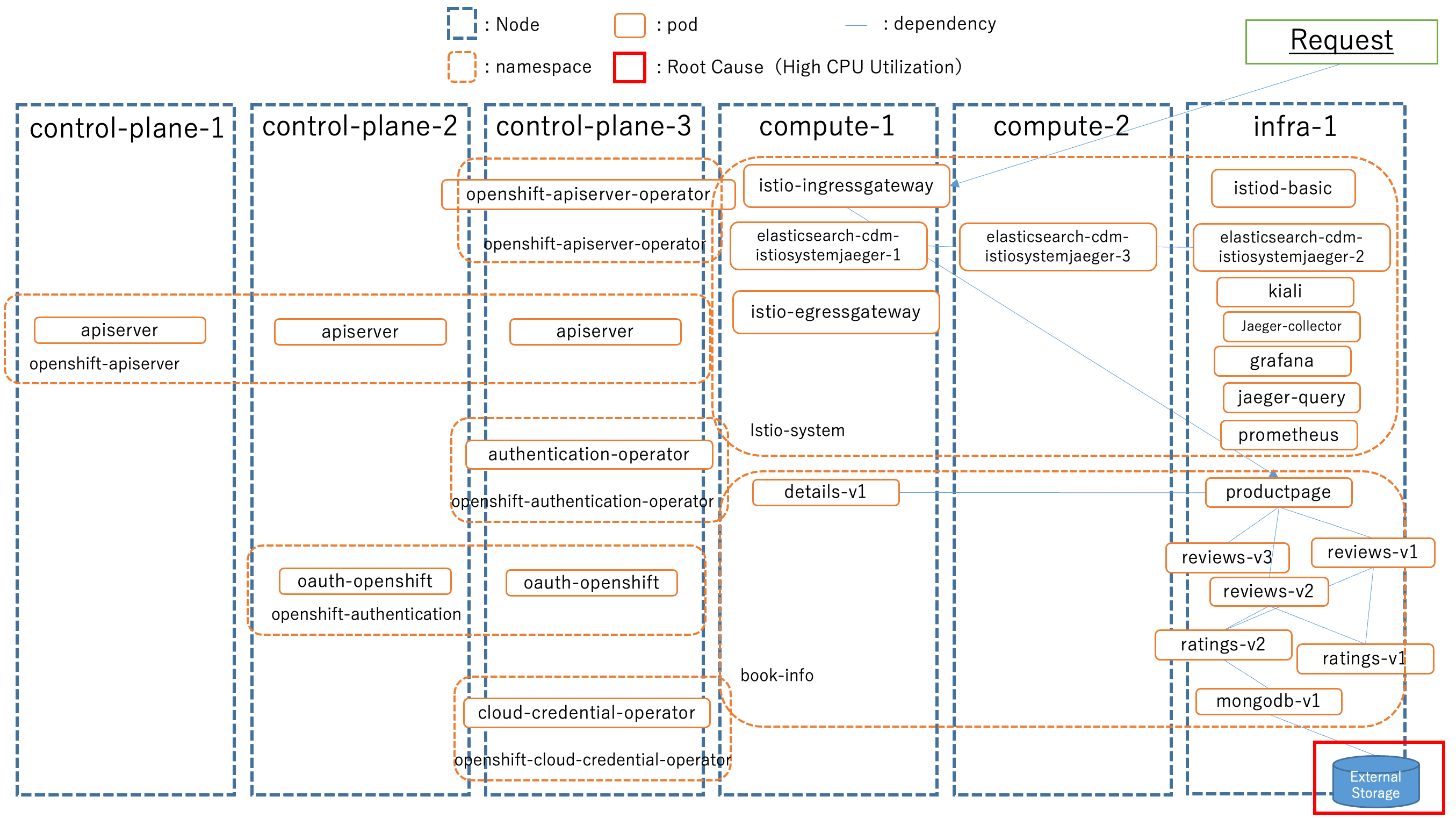}
    \caption{Corresponding to \Cref{fig_aiops_structure} (a). The architecture of Product Review Platform}
    \label{fig:system-structure-apdx}
\end{figure}

\begin{figure}
    \centering
    \includegraphics[width=1\linewidth]{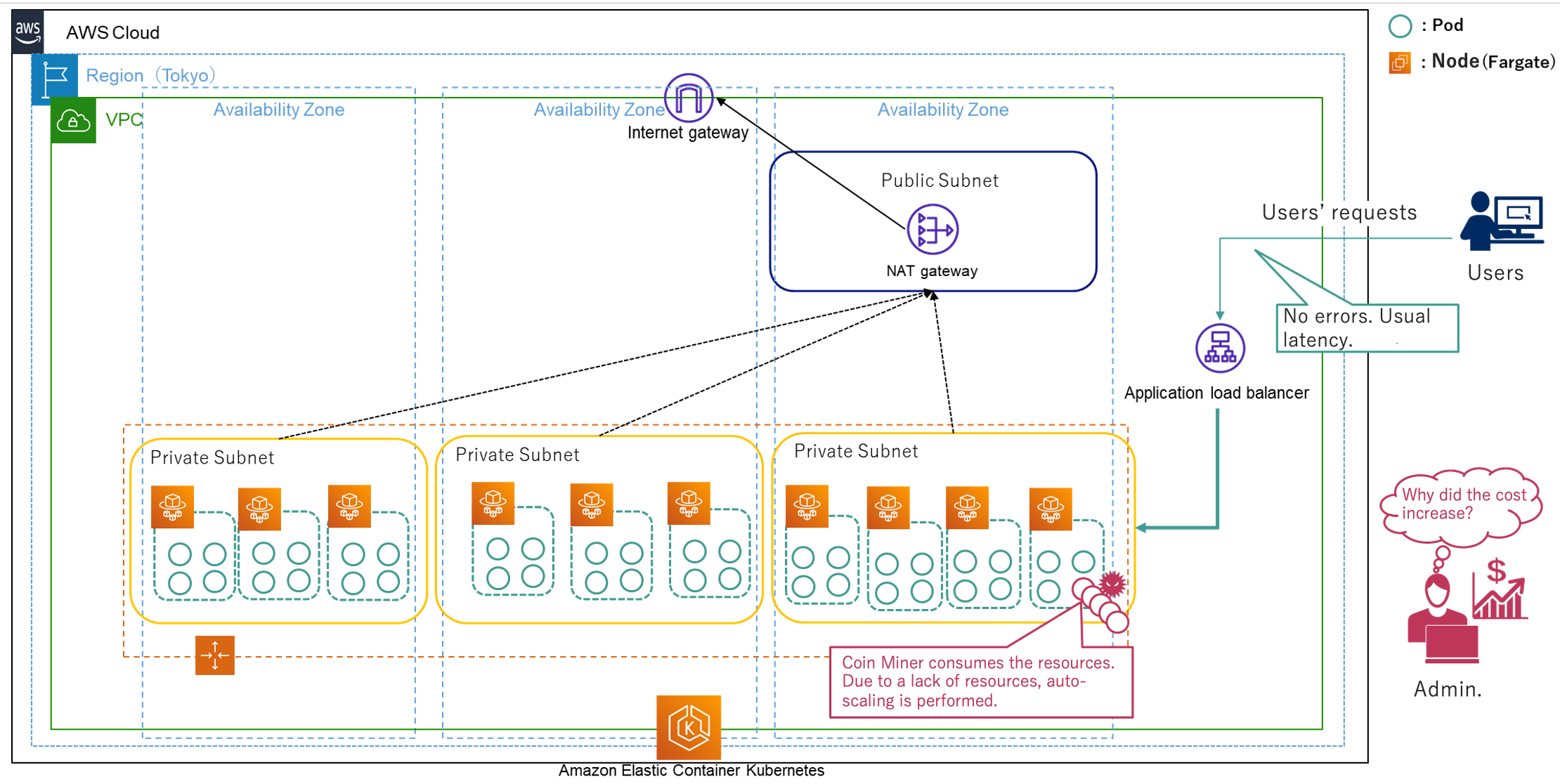}
    \caption{Corresponding to \Cref{fig_aiops_scenarios} left. Visualization of Cryptojacking system fault scenario. \textbf{Right:} External storage failure.}
    \label{fig:crytojacking-apdx}
\end{figure}

\begin{figure}
    \centering
    \includegraphics[width=1\linewidth]{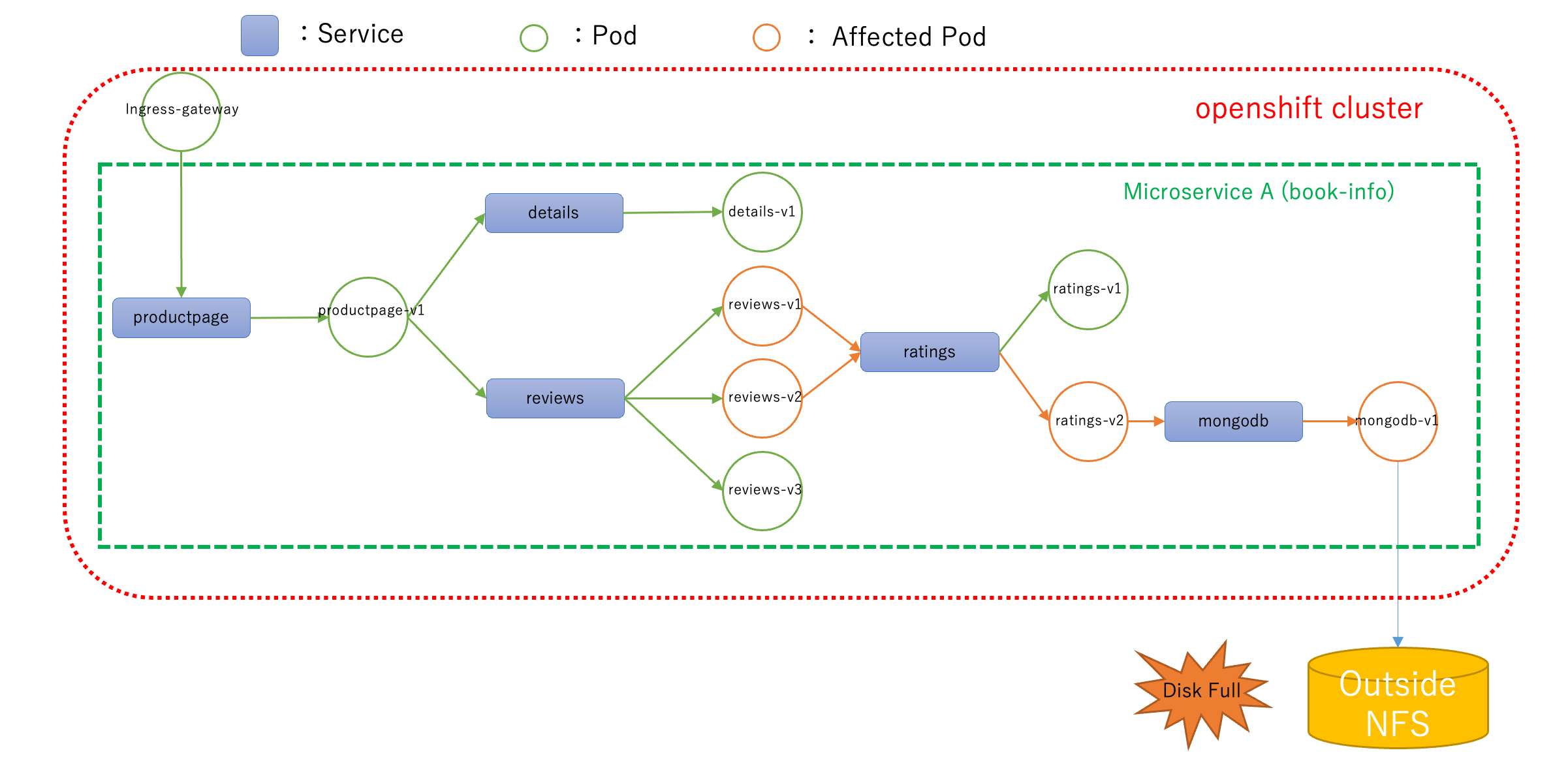}
    \caption{Corresponding to \Cref{fig_aiops_scenarios} right. Visualization of External storage failure. system fault scenario. }
    \label{fig:storage-apdx}
\end{figure}

We provide figures related to the system architecture and fault scenarios in this section, for better readability. The architecture of Product Review Platform is shown in \Cref{fig:system-structure-apdx}, and the system fault scenarios are demonstrated in \Cref{fig:crytojacking-apdx} and \Cref{fig:storage-apdx}.

\section{Dataset Labeling Methodology}
\label{apdx:label-method}

We provide more details on the system fault labeling strategy, which comes in two-fold: the root cause labeling process and label validation. 
\paragraph{Root Cause Labeling Process.}
\begin{itemize}
    \item For each system fault, we designed controlled fault scenarios to mimic realistic fault patterns (e.g., external storage failure, database overload). 
    \item During each controlled fault case, we monitored system behaviors, including metrics and logs, to identify the exact root cause of the fault. 
    \item The ground truth root cause was then labeled based on the specific fault of the system. This ensures high accuracy in root cause labeling, as the faults were systematically induced and their impacts directly observed. 
\end{itemize}

\paragraph{Label Validation.}
\begin{itemize}
    \item To ensure label correctness, we validated the root cause labels by analyzing the system's behavior during and after fault. This involved cross-checking the observed anomalies in system metrics and logs with the expected outcomes of the fault.
    \item Multiple experts reviewed the labeled faults to confirm the consistency and correctness of the root cause assignments.
\end{itemize}

\section{Parameter Settings for All Baseline Models}

We provide the detailed parameter settings for all baseline models as follows:

\begin{itemize}
  \nop{  \item \textbf{Dynotears:} 
    $\texttt{lag=20}$ (\textit{maximum time lags}), $\texttt{lambda\_w=1e-3}$ (\textit{weight regularization}), $\texttt{lambda\_a=1e-3}$ (\textit{autoregressive term regularization}), $\texttt{g\_thre=0.3}$ (\textit{sparsity threshold}).}
    
    \item \textbf{PC:} 
    $\texttt{alpha=0.05}$ (\textit{significance level for conditional independence tests}), $\texttt{ci\_test='fisherz'}$ (\textit{type of conditional independence test}).
    
    \item \textbf{RCD:}
    $\texttt{ci\_test='chisq'}$ (\textit{type of conditional independence test}).
    $\texttt{k=10}$ (\textit{top-k root causes}), 
    $\texttt{alpha\_limit=0.5}$ (\textit{the maximum alpha for search}), 

    \item \textbf{$\varepsilon$-Diagnosis:}
    $\texttt{test\_size=0.6}$ (\textit{train test split ratio}), 
    $\texttt{root\_cause\_top\_k=10}$ (\textit{top-k root causes}), 

    \item \textbf{Circa:}
    $\texttt{test\_size=0.6}$ (\textit{train test split ratio}), 
    $\texttt{root\_cause\_top\_k=10}$ (\textit{top-k root causes}),

    \item \textbf{Baro:} Parameters of this algorithm are related to the Bayesian online change point detection.
    $\texttt{r=50}$ (\textit{magnitude of the hazard function for Bayesian online learning}), 
    $\texttt{k=3}$ (\textit{number of standard deviations from the mean to consider as an anomaly}), 

    \item \textbf{Nezha:} level=service (detection at the service level)
    
 \nop{  \item \textbf{C-LSTM:} 
    $\texttt{hidden=100}$ (\textit{hidden units in LSTM}), $\texttt{lag=20}$ (\textit{maximum time lags for sequence modeling}), $\texttt{lam=10.0}$ (\textit{model complexity regularization}), $\texttt{lam\_ridge=1e-2}$ (\textit{ridge regression regularization}), $\texttt{lr=1e-3}$ (\textit{learning rate}), $\texttt{max\_iter=30000}$ (\textit{maximum iterations}), $\texttt{g\_thre=0.3}$ (\textit{sparsity threshold}).
    
    \item \textbf{GOLEM:} 
    $\texttt{lambda\_1=2e-2}$ (\textit{weight for sparsity regularization}), $\texttt{lambda\_2=5.0}$ (\textit{weight for smoothness regularization}), $\texttt{learning\_rate=1e-3}$ (\textit{optimization learning rate}), $\texttt{num\_iter=30000}$ (\textit{number of iterations for training}), $\texttt{g\_thre=0.3}$ (\textit{sparsity threshold}).
    
    \item \textbf{REASON:} 
    $\texttt{lag=20}$ (\textit{maximum time lags for causal modeling}), $\texttt{L=150}$ (\textit{hidden layers with 150 units}), $\texttt{lambda1=1}$ (\textit{adjacency matrix sparsity regularization}), $\texttt{lambda2=1e-2}$ (\textit{autoregressive term balancing regularization}), $\texttt{gamma=0.8}$ (\textit{integration of individual and topological causal effects}), $\texttt{g\_thre=0.3}$ (\textit{sparsity threshold}).}

\end{itemize}

\section{Dataset Representativeness}
In this section, we aim to show the representativeness of the released dataset. While it is challenging to establish a universal metric for representativeness in benchmarks, we have made significant efforts to ensure the dataset covers diverse fault scenarios:
\begin{itemize}
    \item \textbf{Real-World Fault Scenarios:} The IT domain datasets (Product Review and Cloud Computing) encompass realistic microservice faults such as out-of-memory errors, DDoS attacks, and cryptojacking, as outlined in Section 3.1 and Appendix B. Similarly, the OT domain datasets (SWaT and WADI) include real-world cyber-physical system faults recorded in controlled environments.
    \item \textbf{Diversity of Fault Types:} Across IT and OT domains, we include 10 distinct fault types, ensuring coverage of both transient and persistent system failures. This diversity reflects common issues faced by modern IT and OT systems.
    \item \textbf{Comparative Analysis:} As seen in Table 3 and related discussions, our dataset exhibits performance trends consistent with other benchmarks (e.g., Petshop), supporting its credibility as a representative evaluation platform.
    \item \textbf{Quality Assurance:} All data were collected using industry-standard monitoring tools like Prometheus, CloudWatch, and Elasticsearch. Each fault scenario was validated to ensure it mirrors real-world conditions.
\end{itemize}

\end{document}